%% file: emnlp2021.tex
\pdfoutput=1
\documentclass[11pt]{article}

\usepackage{emnlp2021}

\usepackage{times}
\usepackage{latexsym}
\usepackage{booktabs}
\usepackage{scalerel}
\usepackage{amsmath}
\usepackage{amssymb}
\usepackage{multirow}
\usepackage{refcount}
\usepackage{afterpage}

\usepackage[T1]{fontenc}
\usepackage[utf8]{inputenc}

\usepackage{microtype}

\usepackage{changepage}
\newif\ifcomment
\commenttrue % comment out to disable comments...
\ifcomment
\newcommand{\micomment}[1]{\textcolor{red}{\bf \small [ #1 --MI]}}
\newcommand{\mkcomment}[1]{\textcolor{green}{\bf \small [ #1 --MK]}}
\newcommand{\nacomment}[1]{\textcolor{blue}{\bf \small [ #1 --NA]}}
\else
\newcommand{\micomment}[1]{}
\newcommand{\mkcomment}[1]{}
\newcommand{\nacomment}[1]{}
\fi

\title{The Perils of Using Mechanical Turk\\ to Evaluate Open-Ended Text Generation}

\author{
Marzena Karpinska \hspace{4mm} Nader Akoury \hspace{4mm} Mohit Iyyer \\
University of Massachusetts Amherst\\
  {\tt \{mkarpinska,nsa,miyyer\}@cs.umass.edu}
  }

\begin{document}
\maketitle

\input{sections/abstract}
\input{sections/introduction}
\input{sections/survey}
\input{sections/crowd_experiments}
\input{sections/teacher_experiments}
\input{sections/related_work}
\input{sections/recommendations}
\input{sections/ethics}
\input{sections/acknowledgments}

\bibliography{bib/anthology,bib/custom}
\bibliographystyle{bib/acl_natbib}

\setcounter{table}{0} \renewcommand{\thetable}{A\arabic{table}}
\setcounter{figure}{0} \renewcommand{\thefigure}{A\arabic{figure}}
\setcounter{footnote}{0} \renewcommand{\thefootnote}{\arabic{footnote}}

% Moved here to make the table float on the first page of the appendix
\begin{table*}[!tb]
\begin{tabular}{lllll}
\textbf{Qualifications} & & & & \\ \\
Approval Rate & 90\% (\emph{4})                    & 95\% (\emph{3})                 & 97\% (\emph{1})           & 99\% (\emph{1})    \\
Accepted HITs & 500 (\emph{1})                     & 1000 (\emph{5})                 & 5000 (\emph{1})           &             \\
English       & Resident (6)                & Native Speaker (\emph{4})       & Self-reported (\emph{1})  &             \\
Other         & \multicolumn{2}{l}{Complete all ratings (\emph{1})}    & \multicolumn{2}{l}{Passed quality check (\emph{1})}                            \\
\midrule \\
\textbf{Payment}                       &               \multicolumn{2}{l}{\$0.05 - \$2 per HIT (\emph{8})}     & \multicolumn{2}{l}{\$12 - \$20 per hour (\emph{2})}                                 \\
\midrule \\
\textbf{Number of Items}               &               < 100 items (\emph{9})            & 100 items (\emph{14})           & > 100 items (\emph{20}) &   not reported (\emph{5})         \\
\midrule \\
\textbf{Number of Raters}              &               < 3 raters (\emph{3})                 & 3 raters (\emph{16})                 & 5 raters (\emph{13})           & > 5 raters (\emph{3}) \\
\midrule \\
\textbf{Likert Scale}                  &               3-point (\emph{4})                     & 4-point (\emph{3})                  & 5-point (\emph{24})            & 6-point (\emph{1})     \\
\midrule \\
\textbf{Ranking Task}                  &               two texts (\emph{19})                   & more texts (\emph{4})              &                    &             \\
\midrule \\
\textbf{Text Length}                   &               sentence (\emph{28})                     & paragraph (\emph{21})                &                    & \\
\end{tabular}
\caption{Results of the survey from Section \ref{sec:survey}. Numbers in brackets refer to the number of papers/experiments which employed the given measure.}
\label{tab:survey_table}
\end{table*}

\newpage
~
%\newpage

\appendix
\input{sections/appendixa}
\input{sections/appendixb}
\input{sections/appendixc}
\input{sections/appendixd}

\end{document}

%% file: sections/abstract.tex
\begin{abstract}

Recent text generation research has increasingly focused on open-ended domains such as story and poetry generation. Because models built for such tasks are difficult to evaluate automatically, most researchers in the space justify their modeling choices by collecting crowdsourced human judgments of text quality (e.g., Likert scores of coherence or grammaticality) from Amazon Mechanical Turk (AMT). In this paper, we first conduct a survey of 45 open-ended text generation papers and find that the vast majority of them fail to report crucial details about their AMT tasks, hindering reproducibility. We then run a series of story evaluation experiments with both AMT workers and English teachers and discover that even with strict qualification filters, AMT workers (unlike teachers) fail to distinguish between model-generated text and human-generated references. We show that AMT worker judgments improve when they are shown model-generated output alongside human-generated references, which enables the workers to better calibrate their ratings. Finally, interviews with the English teachers provide deeper insights into the challenges of the evaluation process, particularly when rating model-generated text. 

% Human evaluation plays a central role in NLP. It is often conducted on crowdsourcing platforms, such as Amazon Mechanical Turk (AMT). While numerous studies reported reliability issues related to crowdsourcing platforms the results of such an evaluation are rarely questioned by researchers which may lead to compromised conclusions. To provide a deeper insight into this issue we conduct a series of evaluations of machine generated and human written texts employing both workers on AMT and professional teachers. We demonstrate that the results of such an evaluation may not be consistent even when workers on AMT are recruited using strict qualification filters. Furthermore, we show that presenting system's output along with human written text facilitates the evaluation leading to more reliable ratings. We also conduct a series of interviews with the teachers to gain deeper insights of the evaluation process and highlight potential challenges. Finally, we conduct a survey of  2018-2020 language generation papers which employ human evaluation showing that this kind of evaluation is often understated with a significant number of papers failing to report crucial details.

\end{abstract}

%% file: sections/introduction.tex
\section{Introduction}
\label{sec:intro}

Recent advances in neural language modeling have spurred research into open-ended text generation tasks such as story generation~\citep{peng2018towards}, style transfer~\citep{krishna-etal-2020-reformulating}, and pun generation \citep{he-etal-2019-pun}. Since the space of possible outputs for these tasks is huge compared to more constrained problems such as machine translation, automatic metrics such as BLEU~\citep{bleu} and ROUGE~\citep{Lin2004ROUGE} that measure similarity to reference texts are mostly uninformative~\citep{akoury-etal-2020-storium}.\footnote{Nevertheless, such metrics are commonly reported in research papers on open-ended text generation.} Human evaluation of model-generated text, which is critical for open-ended tasks given the unreliability of automatic metrics~\citep{Peng2017_automatic_metric,Reiter2018_automatic_metric,see2019massively}, is frequently conducted on Amazon's popular Mechanical Turk platform (AMT) to minimize cost and time. Most existing AMT studies ask crowdworkers to provide Likert scale ratings of various properties of generated text, such as \textit{fluency} and \textit{likability}. 

In this paper, we study the reliability and reproducibility of AMT evaluations of open-ended text generation. We first conduct a survey of papers on open-ended text generation between 2018-2020 and find many critical details often go unreported (e.g., worker qualifications, payment, task descriptions, annotator agreement), a finding in line with prior reproducibility studies outside open-ended text generation~\citep{Card2020WithLP_little_power,howcroft-etal-2020-twenty,vanderLee2021}.

Next, we perform a series of story generation evaluations with both AMT workers and expert raters (English teachers), applying a variant of the most common task configuration that appeared in our survey (5 point Likert scale ratings of 200 examples with three annotators per example) to the paragraph-length \texttt{WritingPrompts} dataset of~\citet{fan2018hierarchical}. Unlike prior work in this area, we ask raters to evaluate both stories generated by a fine-tuned GPT-2 language model~\citep{radfordGPT2} and \emph{human-written reference} stories on the same scale, as we expect the latter to consistently score higher on all evaluations. Our experiments expose and quantify several troubling trends:

\begin{figure*}[t]
    \centering
    \includegraphics[width=\textwidth,height=\textheight,keepaspectratio]{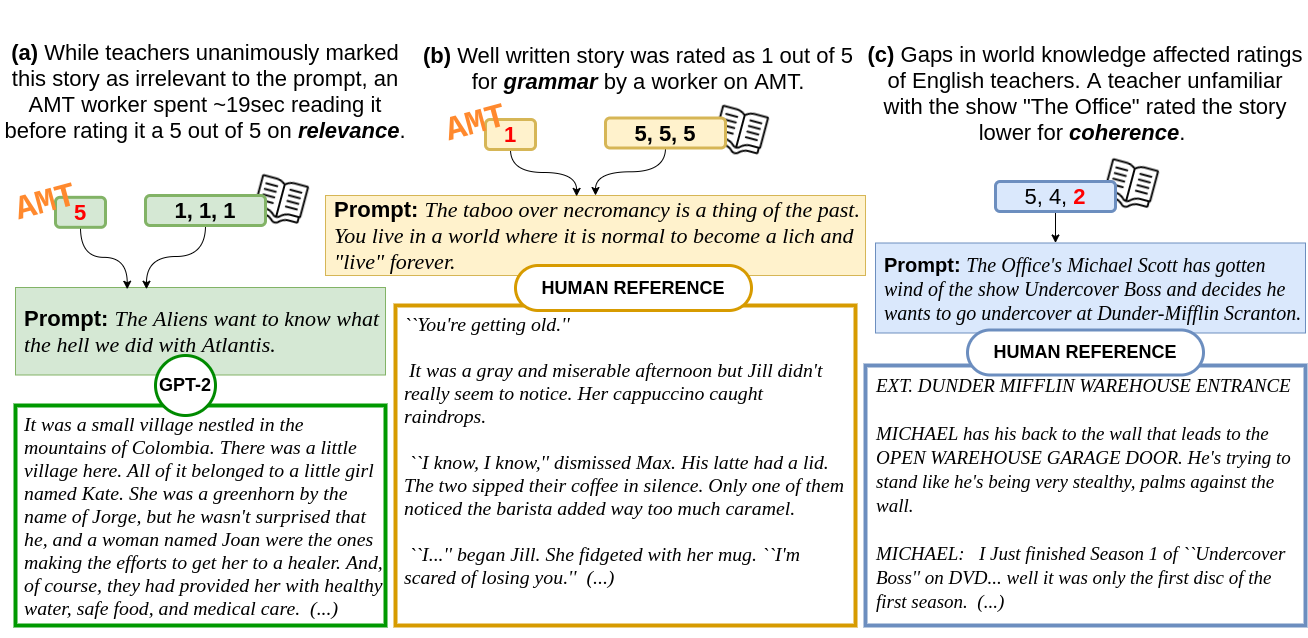}
    \caption{Three examples of prompt-story pairs along with ratings from AMT workers and expert teachers that demonstrate major issues with human evaluation of open-ended text generation.}
    \label{fig:stories_examples}
\end{figure*}

\begin{enumerate}
\setlength{\itemsep}{2pt}
\setlength{\parskip}{0pt}
\setlength{\parsep}{0pt}
    \item AMT ratings do not reliably distinguish model-generated text from human-generated text unless workers are asked to rate both side-by-side, which allows them to better calibrate their ratings.
    \item Running an identical task (same AMT parameters and input data) on different days of the week exhibits high variance and can lead to dubious conclusions (e.g., that reference texts are lower quality than GPT-2 generated text).
    \item Many AMT workers do not carefully read the text that they are evaluating. Even after enabling multiple qualifications to exclude low-quality workers, 42\% of workers on average take fewer than 40 seconds to complete each task. Filtering out these workers can make a significant impact to the overall ratings, but also notably reduces the number of datapoints.
    % \item Judgments provided by AMT workers from non-English speaking countries deviate the most from those of experts, which makes sense given that we are evaluating English text. However, most papers we surveyed do not restrict their worker population to English-speaking countries (or do not report doing so). 
    \item Even expert raters struggle to read and judge model-generated text. The time they spend per example increases significantly compared to that for references, and agreement also drops. 
    % \item \micomment{something else about what is hardest for experts to rate, and also about what they did when they were asked to rate things together?}
\end{enumerate}

For future human evaluations of open-ended text generation tasks, we urge researchers to obtain expert raters whenever possible. If AMT is the only feasible option, we recommend that available reference outputs also be evaluated alongside model-generated ones to improve rating calibration, and also that heavy filtering of the worker population (possibly through qualification tasks, or post-hoc removal) is performed prior to reporting results.

% In the course of our study: (1) we attempt to reproduce a common research setting which shows that results of such an evaluation are not always consistent even if we filter the workers by their approval rate and/or number of completed HITs. (2) We demonstrate that presenting machine generated text along with the human written text facilitates the evaluation task making it easier to compare the two batches. (3) We employ professional teachers showing that they can provide higher quality ratings with higher agreement even if no specific guidelines are provided. (4) We conduct interviews with the teachers in order to get a deeper insight into the evaluation process and techniques they employed. At last, (5) we conduct a survey of 2018-2020 NLP papers which use AMT to evaluate generated text summarizing the current practice. We also show that researchers tend to understate the importance of human evaluation failing to report details about the experimental setting. 

%% file: sections/survey.tex
\section{A survey of papers that evaluate open-ended text generation with AMT}
\label{sec:survey}
We begin with a survey of \emph{45} papers that use AMT to evaluate the output of open-ended English-language text generation models, which includes generated stories, metaphors, paraphrases, puns, sarcasm, and sentences with transferred style or attributes~\footnote{The details on the survey questions and surveyed papers are provided in the \autoref{sec:appendixa_survey}}. Each paper was published between 2018 and 2020 at ACL, NAACL, or EMNLP, and we exclude papers that use AMT to evaluate more well-established generation tasks like machine translation, or summarization.\footnote{For detailed numbers see \autoref{tab:survey_table}; the specific papers in the survey are in \autoref{tab:surveyed_papers}. Four papers conducted multiple evaluations with different settings, and as such we count them in multiple categories. Moreover, all but nine papers evaluated their systems on more than one attribute.} Unlike previous surveys of evaluating generated text~\citep{elikyilmaz2020EvaluationOT,vanderLee2021}, we focus specifically on AMT evaluations of open-ended text generation. 
In this section, we provide an overview of the different types of evaluation task setups present in our survey; later, we experiment with several variants of the most common setup.

% % \mkcomment{uncommented the passsage below}
% % The most commonly used methods of evaluation in our survey are the 5-point Likert scale, followed by choosing between two or more system outputs. The most commonly rated criteria are \textit{fluency} and/or \textit{grammaticality}, \textit{overall quality}, \textit{relevance} and \textit{coherence}. Furthermore, \emph{14} of the surveyed papers fail to report essential details of the experimental setup (i.e., number of raters per item, and/or number of items rated) hindering reproducibility. This is in line with findings reported in \citet{vanderLee2019},
% % \citet{Card2020WithLP_little_power}, \citet{howcroft-etal-2020-twenty} and \citet{vanderLee2021}. In the rest of this section, we analyze the most common task setup, including employed rating scales, number of rated items, and qualification filters.

\paragraph{Evaluation criteria:} As in the survey of~\citet{howcroft-etal-2020-twenty}, we observe a variety of different evaluation criteria and definitions of these criteria across the \emph{45} papers. The most common criteria include \textit{fluency} and/or \textit{grammaticality} (\emph{19}), \textit{overall ``quality''} (\emph{12}), and \textit{relevance} (\emph{10}) to a corresponding prompt. Furthermore, stories in particular tend to also be evaluated on some notion of \textit{coherence} (\emph{9}) and \textit{likability} (\emph{4}).

\paragraph{Rating scales:} 
\label{rating_scale_desc}
More than half of the papers (\emph{24}) employ a 5-point Likert scale to evaluate the above criteria; of these, \emph{19} provided labels for just the end points of the scale (e.g., ``lowest'' vs. ``highest''), while \emph{5} labeled all points on the scale. The next most common evaluation type is ranking two or more system outputs (\emph{23}). Less common are other Likert scales (3, 4, or 6-point), pass/fail tasks, and output-prompt matching tasks.

\paragraph{Number of raters and rated items:} An alarming number of papers (\emph{14}) do not even report the number of raters 
and/or items (\emph{6})  
used for evaluation. Of the remaining, most papers (\emph{16}) obtain ratings from 3 separate AMT workers per item. The most common number of items per evaluation is 100 (\emph{14}). The number of raters per item in other studies ranges from 2 to 11, while the number of items ranges from 12 to 1,000.

\paragraph{Workers qualifications and compensation:}
The vast majority of papers do not report AMT worker qualifications (\emph{32}) or worker compensation (\emph{35}), which adds to the reproducibility woes. Among papers that report qualifications, the most common were HIT\footnote{In AMT parlance, a human intelligence task (HIT) refers to a single item; in our case, each HIT corresponds to one story, which workers rate on four different properties.} approval rate $\geq$ 90\%-99\%, and number of approved HITs between 500 to 5,000. Only \emph{11} papers mentioned restricting workers to those from English-speaking countries or applying some kind of language test, despite all evaluations being done on English text.

\paragraph{Length of the Rated Text:} 
As open-ended text generation encompasses an array of different tasks, the length of the rated text differed greatly, ranging from single sentences (\emph{28}), sometimes presented in a longer context, to short paragraphs (\emph{7}), and longer paragraphs (\emph{14}). The latter setting is most commonly used for story generation tasks.

% \paragraph{Reported Instructions}
% Only 11 papers provided details on the exact instructions given to the workers, while 22 provided only a general overview (see \autoref{tab:survey_table}).\micomment{this feels a little vague (what is detailed vs general overview, commetned out for now}

%summary paragraph highlighting the most common ways in whcih MTurk eval tasks are setup

%what are the types of tasks that were included in survey (not translation / summarization). mostly story generation and dialogue.      
%table with papers and different categories

% \paragraph{}
% For our experiments, we employ the story generation task where the system has to generate a story based on the provided prompt. 
% %While our survey revealed that most studies evaluated 100 items, we decided to increase this number to 200 items since 
% We elicitated ratings on four attributes commonly associated with this task, that is \textit{grammar}, \textit{coherence}, \textit{relevance} (to the prompt), and \textit{likability} on a 5-point Likert scale where only the end points are labelled.

%For our experiments, we analyze the most common X properties of these these tasks to assess their reliability.

%% file: sections/crowd_experiments.tex
\section{Evaluating story generation with AMT}
\label{sec:crowd_experiments}

Our survey reveals that the most popular Mechanical Turk task design for open-ended text generation asks AMT workers to rate various properties of generated text on a 5-point Likert scale. In this section, we conduct a series of AMT evaluations for the open-ended problem of \emph{story generation} by varying different parameters within this standard task design. Importantly, we evaluate both model-generated stories as well as \emph{human-generated reference} stories, which provides a pseudo upper bound for the ratings. Our experiments reveal that worker qualifications (e.g., HIT approval rate and number of accepted HITs) do not notably impact judgments or spam rate on reference stories, with the exception of country of origin.
Furthermore, we uncover an issue with \emph{rating calibration}: when both reference and model-generated stories are included for the same prompt, average reference scores are significantly higher than those for model-generated text; however, when workers only see one type of text per HIT, they give similar average scores to both types. 
% We conduct a series of experiments on AMT evaluating both stories written by humans and stories generated by GPT-2 which showed that the two factors, that is approval rate and number of accepted HITs, 
% %may not be
% are not sufficient to effectively filter out workers who are likely to spam the task.

\subsection{Experimental Setup}
\label{sec:crowd_experimental_setup}

We first describe the parameters of our experiments before later analyzing the results. 

\paragraph{Dataset:} We use the \texttt{WritingPrompts} dataset collected by~\citet{fan2018hierarchical}, which is a collection of 303,358 English language stories written by Reddit users on the \texttt{r/WritingPrompts} subreddit.\footnote{\url{https://reddit.com/r/WritingPrompts/}} This dataset, which consists of short prompts  paired with user-written stories (e.g., \href{https://www.reddit.com/r/WritingPrompts/comments/j371rg/wp_there_are_10_legendary_dentists_who_review/}{``There are 10 legendary dentists who review every toothpaste. You are the 10th... being hunted by the other 9...''}), has been used in multiple previous works on paragraph-length story generation \citep{fan-etal-2019-strategies, see2019massively, mao-etal-2019-improving}. We randomly select 200 prompts from the test set for all of our experiments. Since the human-written stories in the dataset are already tokenized, we first de-tokenized the stories, cleaned up artifacts from lemmatization, and manually truncated each story so that it ends with a full sentence and is no longer than 150 words in order to make the length comparable with the machine-generated story.\footnote{The mean length of the selected reference stories is 134.5 tokens, with a standard deviation of 8.81.} We use the resulting stories for all experiments with reference text.

\paragraph{Model-generated stories:}
We follow a similar modeling approach to prior story generation work~\citep{mao-etal-2019-improving,guan-etal-2020-knowledge} by fine-tuning a pretrained GPT-2 medium-sized model~\citep{radfordGPT2} on the training set of the \textit{WritingPrompts} dataset, using the HuggingFace Transformers library \cite{wolf-etal-2020-transformers}. We use a batch size of approximately 50$k$ tokens, a learning rate of $5e-5$ with a linear learning rate schedule, and train for 3 epochs, stopping training after validation perplexity converges to $\sim 19$. Each training example consists of a concatenation of a prompt, separator token (new line character), and reference story. At test-time, we feed the same 200 prompts selected above to our model for fair comparison to the human-written stories, and we generate three stories per prompt using nucleus sampling~\citep{holtzman2019curious} with $p=0.9$. We manually truncate each sample so that it ends with a full sentence and is no longer than 150 words.\footnote{The mean length of the generated stories is 137.4 tokens, with a standard deviation of 8.36.} These stories are used in all experiments evaluating machine-generated stories.

\paragraph{AMT task parameters:}
We conduct all experiments using the default interface in Mechanical Turk (see \autoref{fig:one_story_eval} and \autoref{fig:two_stories_eval}). Workers were asked to rate human-written and/or machine-generated stories on four attributes, with the following definitions provided to them: 
\begin{enumerate}
\setlength{\itemsep}{2pt}
\setlength{\parskip}{0pt}
\setlength{\parsep}{0pt}
\item \textit{Grammar}: ``How grammatically correct is the text of the story fragment?''
\item \textit{Coherence}: ``How well do the sentences in the story fragment fit together?''
\item \textit{Likability}: ``How enjoyable do you find the story fragment?''
\item \textit{Relevance}: ``How relevant is the story fragment to the prompt?''
\end{enumerate}

Their ratings fall on a 5-point Likert scale with the corresponding endpoints labelled as ``lowest'' (1 point) and ``highest'' (5 points). Since our survey did not find many previous papers that reported using detailed descriptions for each point on the scale, we chose to use minimal labels to mimic the most popular setup (see \autoref{rating_scale_desc} for details). 

Each of our AMT experiments shows workers the same 200 prompts paired with human and/or machine-generated stories, and we solicit three worker judgments per HIT. Workers were paid \$0.20 per HIT for tasks that showed one story, and \$0.35 per HIT for those that showed two stories; in total, our AMT experiments cost roughly \$1.5K. Importantly, each experiment used a completely different set of workers (i.e., each worker could only participate in one experiment, although they can complete multiple HITs within that experiment), which is an intentional choice to prevent workers from judging the same story multiple times. Finally, to eliminate potential variations stemming from evaluation on different days (weekdays vs. weekends) and time of day, we launch all experiments on weekdays between 11:00-11:30AM PST.
% All experiments were conducted using the AMT interface. Except for Experiment 9 where two stories were presented in one HIT, workers were asked to rate one story at a time on four attributes: (1) \textit{grammaticality} (``How grammatically correct is the text of the story fragment?''), (2) \textit{coherence} (``How well do the sentences in the story fragment fit together?''), (3) \textit{likability} (``How enjoyable do you find the story fragment?''), and (4) \textit{relevance} (``How relevant is the story fragment to the prompt?'') on a 5-point Likert scale with the corresponding endpoints labelled as ``lowest'' and ``highest'' (see [APPENDIX] for details). 

% The following paragraphs describe MTurk experiments in more detail. Except for experiment 9 which included a total of 400 stories, 200 written by humans and 200 generated by machine, in all the experiments workers were asked to evaluate 200 stories, either written by humans or machine-generated. Each story was evaluated by 3 workers. Workers were allowed to participate only in one experiment, however, they could provide ratings for multiple stories by completing multiple HITs. [TABLE] summarizes setups of all MTurk experiments.

\subsection{AMT Evaluations of Reference Text}
Our first set of experiments concerns only human-written reference stories; we move to machine-generated text in the next subsection. One of our assumptions with human-written stories, supported by the expert teacher assessment in \autoref{sec:teacher_experiments}, is that they should receive relatively high scores for all four properties (except perhaps \emph{likability} which is highly subjective). We thus use reference texts to evaluate various AMT parameters such as qualifications or day of task launch, observing how modifications to these parameters affect the average scores of reference text. 

% We also employ these stories to assess the importance of qualifications applied via MTurk interface (i.e., approval rate, number of approved HITs) %, and workers` country of origin) 
% as well as the workers` country of origin. Furthermore, we apply the same qualifications to evaluate the same stories in order to estimate the inter-day variance.  

\begin{table*}[ht]

    \small
    \centering
    \begin{tabular}{lllllllll}
    \toprule
    \multirow{2}*{\textbf{Experiment description}}  & \multicolumn{2}{c}{\hspace*{-3ex}\underline{\textbf{Grammar}}} & \multicolumn{2}{c}{\hspace*{-3ex}\underline{\textbf{Coherence}}} & \multicolumn{2}{c}{\hspace*{-3ex}\underline{\textbf{Relevance}}} & \multicolumn{2}{c}{\hspace*{-3ex}\underline{\textbf{Likability}}} \\
    & Mean$_{\text{\emph{STD}}}$ & IAA$_{\text{\emph{\%}}}$ & Mean$_{\text{\emph{STD}}}$ & IAA$_{\text{\emph{\%}}}$ & Mean$_{\text{\emph{STD}}}$ & IAA$_{\text{\emph{\%}}}$ & Mean$_{\text{\emph{STD}}}$ & IAA$_{\text{\emph{\%}}}$ \\
    \midrule
    & \multicolumn{8}{c}{\emph{Impact of Qualifications}} \vspace{0.15cm}\\
    No qualifications & 4.05$_{{0.90}}$ & 0.08$_{{14.5}}$ & 3.92$_{{0.98}}$ & 0.02$_{{4.5}}$ & 3.66$_{{1.22}}$ & 0.13$_{{11}}$ & 3.64$_{{1.16}}$ & 0.02$_{{7}}$ \\
    + > 90\% HIT approval & 4.16$_{{0.86}}$ & 0.07$_{{18}}$ & 4.07$_{{0.93}}$ & 0.06$_{{10.5}}$ & 3.67$_{{1.14}}$ & 0.07$_{{9}}$ & 3.68$_{{1.15}}$ & 0.08$_{{10}}$ \\ 
    + at least 1000 HITs & 3.91$_{{0.85}}$ & 0.05$_{{12}}$ & 3.85$_{{0.98}}$ & 0.08$_{{11.5}}$ & 3.60$_{{1.15}}$ & 0.18$_{{8}}$ & 3.63$_{{1.13}}$ & 0.07$_{{12.5}}$ \\
    + English-speaking countries & 4.00$_{{0.92}}$ & 0.21$_{{15.5}}$ & 4.11$_{{0.96}}$ & 0.14$_{{16.5}}$ & 3.71$_{{1.26}}$ & 0.27$_{{10}}$ & 3.37$_{{1.18}}$ & 0.11$_{{7.5}}$ \\
    \midrule
    & \multicolumn{8}{c}{\emph{Variance Across Days}} \vspace{0.15cm}\\
    Day 1 (all quals.) & 4.00$_{{0.92}}$ & 0.21$_{{15.5}}$ & 4.11$_{{0.96}}$ & 0.14$_{{16.5}}$ & 3.71$_{{1.26}}$ & 0.27$_{{10}}$ & 3.37$_{{1.18}}$ & 0.11$_{{7.5}}$ \\
    Day 2 (all quals.) & 3.86$_{{0.92}}$ & \parbox[b][\baselineskip]{0pt}{\hspace*{-0.9ex}-}0.03$_{{10.5}}$ & 3.92$_{{0.98}}$ & \parbox[b][\baselineskip]{0pt}{\hspace*{-0.9ex}-}0.03$_{{6.5}}$ & 3.71$_{{1.08}}$ & 0.02$_{{11}}$ & 3.73$_{{0.97}}$ & \parbox[b][\baselineskip]{0pt}{\hspace*{-0.9ex}-}0.04$_{{8.5}}$ \\
    Day 3 (all quals.) & 3.98$_{{0.96}}$ & 0.18$_{{11}}$ & 4.05$_{{0.94}}$ & 0.13$_{{10.5}}$ & 3.46$_{{1.29}}$ & 0.26$_{{8}}$ & 3.42$_{{1.16}}$ & 0.07$_{{4.5}}$ \\
    \midrule 
    & \multicolumn{8}{c}{\emph{Impact of Country of Origin}} \vspace{0.15cm}\\
    - English-speaking countries & 3.82$_{{1.04}}$ & 0.03$_{{11}}$ & 3.45$_{{1.19}}$ & \parbox[b][\baselineskip]{0pt}{\hspace*{-0.9ex}-}0.01$_{{9}}$ & 3.25$_{{1.27}}$ & 0.03$_{{6.5}}$ & 3.32$_{{1.26}}$ & \parbox[b][\baselineskip]{0pt}{\hspace*{-0.9ex}-}0.09$_{{3}}$ \\
    \midrule 
    & \multicolumn{8}{c}{\emph{Impact of Filtering by the Median Work Time}} \vspace{0.15cm}\\
    Day 2 (Median $\geq$ 40s) & 4.04$_{{0.94}}$ && 4.33$_{{0.92}}$ && 3.74$_{{1.34}}$ && 3.67$_{{1.06}}$ \\
    
    \bottomrule
    \end{tabular}
    \caption{AMT experiments on human-written reference stories. Inter-annotator agreement (IAA) between the three raters is measured with Krippendorff's $\alpha$ as well as the percentage of stories for which all three raters exactly agreed on a rating (the latter is subscripted). Last section omits IAA due to the large number of missing datapoints. Statistical significance for the relations between groups is provided in \autoref{sec:appendixd_mturk_results}.}
    \label{tab:AMT_only_human_stories}
\end{table*}

%Evaluating the reference text: why did we do this, we expect there to be higher scores on average since these are human-written. figure out how best to talk about agreement ---> probably need another section, this is getting too complicated

\paragraph{Impact of worker qualifications:}
We run four experiments evaluating the previously-described set of 200 prompts with reference story fragments, varying the worker qualifications as follows: (1) no qualifications, (2) including only workers with HIT approval rate > 90\%, (3) including only workers with approval rate > 90\% and at least 1000 approved HITs, (4) including only workers with approval rate > 90\% and at least 1000 approved HITs who are located in English-speaking countries.\footnote{We include workers from the US, Canada, the UK, Australia, and New Zealand.} 

The results in the top portion of~\autoref{tab:AMT_only_human_stories} suggest that applying all of the qualifications (i.e., workers from English-speaking countries, approval rate > 90\%, approved HITs $\geq$ 1000) has a positive effect on the quality of workers, as this setting yielded the highest %reasonably
scores out of the four experiments for \textit{coherence} and \textit{relevance} while ratings for \textit{grammar} were also considerably high. Ratings for \textit{likability} were lower than in the experiments with less strict qualifications, but \textit{likability} is a very subjective measure which consistently shows a very low agreement (Krippendorff's $\alpha$ of -0.04 to 0.11).   %should this be described in more details? 
When all AMT worker qualifications are enabled, the worker ratings more closely align to those made by English teachers, although there are still substantial deviations (\autoref{sec:teacher_experiments}). Additionally, with all qualifications enabled, workers show higher agreement for \textit{grammar}, \textit{coherence}, \textit{relevance} and even \textit{likability}, although the agreement between raters remains low.

%To investigate this issue further, we run the same evaluation experiment applying all the qualification filters (i.e., approval rate > 90\%, number of approved HITs >= 1000, and location being an English speaking country) on two different days. We found 

\paragraph{High variance across different days:}
Concerned by the low overall agreement, we decided to run another set of experiments that repeats the same experiment (all qualifications enabled) across three different days. Due to our constraint that each worker can only participate in one experiment, each of these experiments has a different subset of qualified workers. As shown in the second portion of~\autoref{tab:AMT_only_human_stories}, although the first and third days yielded similar mean ratings/agreement in terms of \textit{grammar} (M=4.00, IAA=0.21 vs M=3.98, IAA=0.18) and \textit{coherence} (M=4.11, IAA=0.14 vs M=4.05, IAA=0.13), the second day received lower ratings across the board and had overall poor IAA (see \autoref{tab:AMT_only_human_stories}). Furthermore, ratings for \textit{relevance} in the third day (M=3.46) 
were significantly lower %(\textit{t(1197.6)=3.44, \textit{p}<0.001})
than in the first two days (M=3.71), which indicates that simply using all AMT qualifications is not enough to achieve consistent results.

\begin{figure}[t]
    \centering
    \includegraphics[width=\columnwidth]{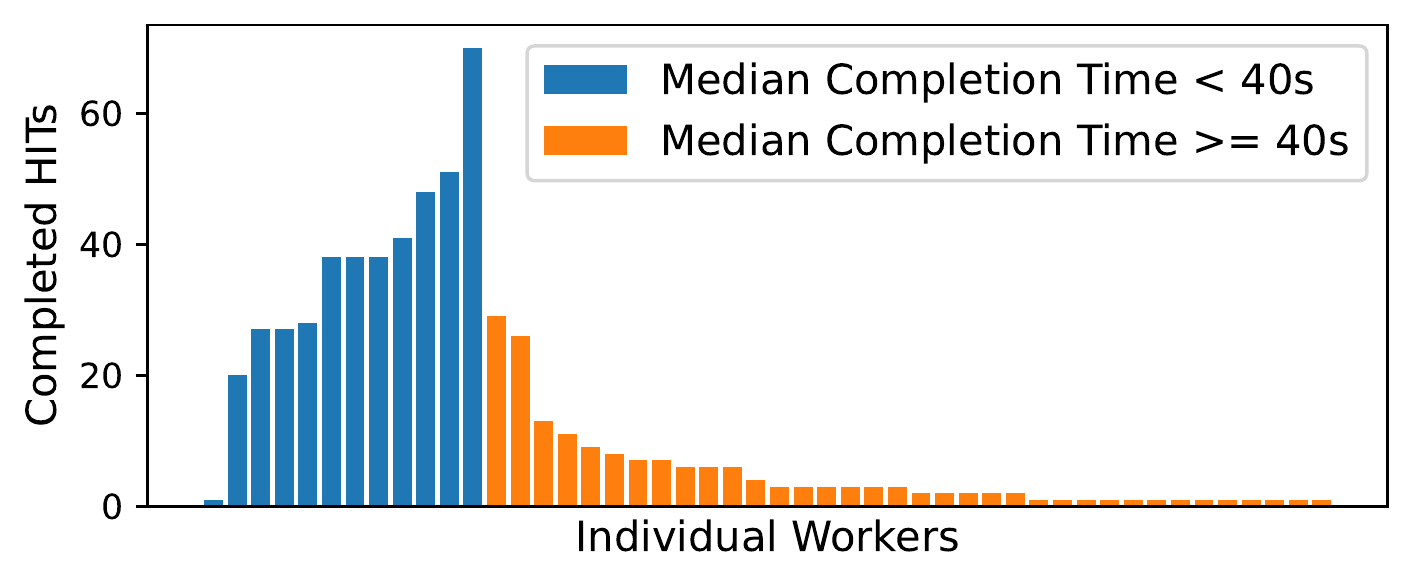}
    \caption{This example from Day 2 shows that many AMT workers complete multiple HITs in less time and with lower agreement in comparison to our experts.}
    \label{fig:mturk_analysis}
\end{figure}

\paragraph{Many AMT workers do not spend enough time reading the stories:}
The low overall agreement also motivated us to examine the average time each worker spent per HIT. While AMT reports \textit{WorkTimeInSeconds} in the results file made available to task requesters, we observe similar to~\citet{akoury-etal-2020-storium} that these times are artificially inflated due to workers who accept multiple HITs at the same time and work on them sequentially (e.g., in different tabs). Such workers are also frequently among the most prolific in terms of HITs completed per experiment (see \autoref{fig:mturk_analysis}), since there is no maximum number of HITs per worker.\footnote{Like most AMT tasks~\citep{Fort2011_prod_workers, Robinson2019_superworkers}, the majority of HITs for our evaluations are provided by a small fraction of workers. The majority of workers provided ratings for only one or two stories while a very few productive workers rated over 50\% of the stories (see \autoref{fig:mturk_analysis}).} We correct for this by measuring the time between consecutively submitted HITs by the same worker, which can be derived by analyzing start and end times of each HIT. This ``actual time'' differs considerably from the AMT reported \textit{WorkTimeInSeconds}: for instance, a worker that AMT reports had a mean work-time of 360 seconds had an actual mean working time\footnote{The mean work time is also not very representative as workers typically \href{https://www.reddit.com/r/mturk/comments/igh9g5/avoid\_story\_bot\_ai/g2unm42/}{accept multiple HITs, wait a period of time, then submit all accepted HITs in quick succession}.
%leading to a large difference between mean and median work time (up to 60s) with the median working time being usually half of the mean working time.
}
of 22s and a median of 13s. To put these numbers in perspective, this is about one-fourth of the time that the fastest English teacher achieved (see \autoref{sec:teacher_experiments}). %the for a task that should take at least 30-40s to complete.

As it is impossible to carefully read a paragraph-length story and assess all four properties in as little as 13 seconds, we measure the impact on average ratings when filtering out workers who spend too little time per HIT (last row of \autoref{tab:AMT_only_human_stories}). Specifically, we remove judgments from workers whose median time is below 40s (which is a low bar), and find that on average about 42\% of our ratings are filtered out (ranging from 20\%-72\% across all experiments).\footnote{We also ran experiments with even stricter qualification filters (i.e., acceptance rate  $\geq$ 99\% and at least 10,000 approved tasks), but this made no notable difference to the percentage of data being filtered out (35\%). 
This is most likely due to the fact that most requesters are reluctant to reject HITs regardless of quality, which results in an estimated 95\% of workers having an approval rate of 98\% or above \citep{Matherly2019_mturk_research, Wessling2017MTurkCM_99percent}.} Of our surveyed papers, only~\citet{akoury-etal-2020-storium} report actual work time, demonstrating that this is a major issue in modern AMT evaluations of text quality that most researchers have overlooked. 

%  Since workers with approval rate of 99\% or above were also demonstrated to be the most productive group on AMT [REF], even when running evaluation without any filters it is most likely that the majority of ratings being provided by such workers, however, it may not translate into better quality data. 
% %not sure if to write about workers with >=99% approval being most productive 
% Since the evaluated stories are written in English, we further explore the impact of workers' country of origin.

%This means that while the work time reported in the AMT file seemed acceptable, they were, in fact, spending much less time on completing the task. 

%[median, mean, impact on data, percentage of data,  very few / no existing papers that we surveyed do this quality filtering, it makes a big difference] --> even 99\% filter does not work -> because of positive bias (requesters not rejecting HITs, most population in the upper level and 99\% workers are the most productive ones so in the end approval rate and num of completed tasks makes little difference (even with setting 90\% we were probably getting 99). ---> we also investigate the impact of country of origin 

%accept rate, num items rated. mention somewhere that qualification accept rate of 99 vs 90 doesnt make difference, cite that paper.

\paragraph{Impact of worker country of origin:}

While all of the surveyed papers evaluate only English text, only \emph{11} of them reported using some kind of filtering to ensure that workers have sufficient knowledge of English. The default AMT setting does not filter workers by country of origin, which potentially increases the variance of results depending on the English proficiency of  workers who accept HITs. To measure this, we re-run our experiment with all qualifications, except we restrict the task to only workers from countries that do not primarily speak English (i.e., we \emph{exclude} workers from the US, Canada, UK, Australia, New Zealand, Ireland, and Singapore). The third portion of \autoref{tab:AMT_only_human_stories} shows that workers from non-English speaking countries rated \textit{coherence}, \textit{relevance}, and \textit{grammar}\footnote{There was no significant difference between \textit{grammar} ratings collected from raters from non-English speaking countries and ratings collected on Day 2.}  significantly lower than identically-qualified workers from English-speaking countries (Day 1-3).
% While \textit{likability} is a rather subjective measure and, in general, we observed low agreement on this attribute, one can argue that \textit{coherence} and \textit{relevance} were just too difficult to rate for workers from non-English speaking countries. Furthermore,  there is also a significant difference between \textit{grammar} ratings of workers from non-English speaking countries and the ratings we obtained on Day 1 and Day 3.
Thus, researchers rating English text should restrict their tasks to English-speaking countries, although \citet{kennedy_clifford_burleigh_waggoner_jewell_winter_2020_mturk} find that many workers use Virtual Private Networks (VPNs) to take part in tasks restricted to those in the US. %add comment?    

\begin{table*}[ht]
    \small
    \centering
    \begin{tabular}{llllllllll}
    \toprule
    \textbf{Raters} & \textbf{Type of text} & \multicolumn{2}{c}{\hspace*{-3ex}\underline{\textbf{Grammar}}} &
    \multicolumn{2}{c}{\hspace*{-3ex}\underline{\textbf{Coherence}}} & \multicolumn{2}{c}{\hspace*{-3ex}\underline{\textbf{Relevance}}} & \multicolumn{2}{c}{\hspace*{-3ex}\underline{\textbf{Likability}}} \\
     & & Mean$_{\text{\emph{STD}}}$ & IAA$_{\text{\emph{\%}}}$ & Mean$_{\text{\emph{STD}}}$ & IAA$_{\text{\emph{\%}}}$ & Mean$_{\text{\emph{STD}}}$ & IAA$_{\text{\emph{\%}}}$ & Mean$_{\text{\emph{STD}}}$ & IAA$_{\text{\emph{\%}}}$ \\
    \midrule
    & \multicolumn{9}{c}{\emph{AMT workers fail to effectively distinguish between human written and GPT-2 generated stories}} \vspace{0.15cm}\\
    AMT & Ref. (Day 1) & 4.00$_{{0.92}}$ & 0.21$_{{15.5}}$ & 4.11$_{{0.96}}$ & 0.14$_{{16.5}}$ & 3.71$_{{1.26}}$ & 0.27$_{{10}}$ & 3.37$_{{1.18}}$ & 0.11$_{{7.5}}$ \\
    AMT & Ref. (Day 2) & 3.86$_{{0.92}}$ & \parbox[b][\baselineskip]{0pt}{\hspace*{-0.9ex}-}0.03$_{{10.5}}$ & 3.92$_{{0.98}}$ & \parbox[b][\baselineskip]{0pt}{\hspace*{-0.9ex}-}0.03$_{{6.5}}$ & 3.71$_{{1.08}}$ & 0.02$_{{11}}$ & 3.73$_{{0.97}}$ & \parbox[b][\baselineskip]{0pt}{\hspace*{-0.9ex}-}0.04$_{{8.5}}$ \\
    AMT & Ref. (Day 3) & 3.98$_{{0.96}}$ & 0.18$_{{11}}$ & 4.05$_{{0.94}}$ & 0.13$_{{10.5}}$ & 3.46$_{{1.29}}$ & 0.26$_{{8}}$ & 3.42$_{{1.16}}$ & 0.07$_{{4.5}}$ \\
    AMT & GPT-2 & 3.94$_{{0.93}}$ & 0.11$_{{17.5}}$ & 3.82$_{{1.12}}$ & 0.05$_{{7.5}}$ & 3.44$_{{1.41}}$ & 0.10$_{{7}}$ & 3.42$_{{1.25}}$ & 0.02$_{{4.5}}$ \\
    \midrule
    & \multicolumn{9}{c}{\emph{AMT workers score GPT-2 lower when also presented with reference text}} \vspace{0.15cm}\\
    AMT & Reference & 3.83$_{{0.99}}$ & 0.13$_{{12.5}}$ & 3.83$_{{1.1}}$ & 0.07$_{{8}}$ & 3.49$_{{1.26}}$ & 0.20$_{{8}}$ & 3.48$_{{1.08}}$ & 0.03$_{{6.5}}$ \\
    AMT & GPT-2 & 3.82$_{{0.90}}$ & 0.10$_{{12}}$ & 3.39$_{{1.1}}$ & 0.04$_{{9.5}}$ & 2.70$_{{1.26}}$ & 0.06$_{{6.5}}$ & 2.99$_{{1.14}}$ & \parbox[b][\baselineskip]{0pt}{\hspace*{-0.9ex}-}0.04$_{{4}}$ \\
    \midrule
    & \multicolumn{9}{c}{\emph{Teachers rate GPT-2 generated stories lower than AMT workers}} \vspace{0.15cm}\\
    Teachers & Reference & 4.50$_{{0.83}}$ & 0.19$_{{35.5}}$ & 4.38$_{{0.91}}$ & 0.14$_{{25}}$ & 3.82$_{{1.38}}$ & 0.25$_{{16}}$ & 3.69$_{{1.30}}$ & \parbox[b][\baselineskip]{0pt}{\hspace*{-0.9ex}-}0.01$_{{5}}$ \\
    Teachers & GPT-2 & 4.56$_{{0.62}}$ & 0.00$_{{24.5}}$ & 3.73$_{{1.19}}$ & 0.17$_{{13}}$ & 2.54$_{{1.49}}$ & 0.54$_{{25.5}}$ & 2.96$_{{1.46}}$ & \parbox[b][\baselineskip]{0pt}{\hspace*{-0.9ex}-}0.07$_{{3}}$ \\

    \bottomrule
    \end{tabular}
    \caption{Comparison of AMT workers and expert teachers on both human and machine-generated text. Inter-annotator agreement (IAA) between the three raters is measured with Krippendorff's $\alpha$ as well as the percentage of stories for which all three raters exactly agreed on a rating (the latter is subscripted). Statistical significance for the relations between groups is provided in \autoref{sec:appendixc_teachers} and \autoref{sec:appendixd_mturk_results}.}
    \label{tab:AMT_and_teach_hum_gpt2}
\end{table*}

\subsection{Evaluating Machine-Generated Text}
We now turn to AMT evaluation of machine-generated stories produced by the GPT-2 model described in \autoref{sec:crowd_experimental_setup}. Based on our previous experiments with reference texts, we select the ``all qualifications'' setting (i.e., workers from English-speaking countries, approval rate > 90\%, approved HITs $\geq$ 1000) for all GPT-2 AMT tasks. We study two different conditions: (1) HITs contain a prompt and a GPT-2 generated text, and (2) HITs contain a prompt and both a human-written reference story as well as a GPT-2 generated story. In the latter case, we ask AMT workers to rate both texts on each of the four properties. Overall, we observe that workers cannot effectively distinguish between reference and model-generated stories when they are evaluated separately (in terms of average ratings), but that this distinction emerges clearly when they are presented with both types of stories in the same HIT.

\paragraph{When presented only GPT-2 generated text, AMT worker ratings rate them similarly to reference texts, despite obviously worse quality:}
\label{side_by_side_hum_gpt2}
In our first experiment, we follow the protocol from our experiments with human-written reference stories, showing AMT workers a prompt and a model-generated story and asking them to rate it on the same attributes (\textit{grammar}, \textit{coherence}, \textit{relevance}, and \textit{likability}). The results of this evaluation are presented in the upper row of \autoref{tab:AMT_and_teach_hum_gpt2} along with the three sets of ratings of reference stories obtained with the same ``all qualifications'' setting from before (Days 1-3 in \autoref{tab:AMT_only_human_stories}).

Surprisingly, GPT-2 output is \emph{not} consistently rated significantly lower than human-written text. For instance, workers in Day 2 rated human-written stories similarly to the GPT-2 generated stories in terms of \textit{grammar} (M=3.86 vs. M=3.94) and \textit{coherence} (M=3.92 vs. M=3.82), while workers in Day 3 rated human-written stories as similarly \textit{relevant} to the prompt as GPT-2 output (M=3.46 vs. M=3.44). Depending on which reference day we compare the GPT-2 output to, GPT-2 is rated similarly to human-written stories in terms of all four properties, which indicates that this evaluation is uninformative; nevertheless, the majority of surveyed papers use exactly this task design to obtain ratings for model-generated output. 
% \nacomment{i'd put an exact number of papers from the survey here} \mkcomment{I'm actually not sure what is meant here by "this methodology", rating system outputs only? if so, then 21 papers did not include human reference, BUT this doesn't mean that 24 included them in a meaningful way, I don't have the numbers but I would say that majority of them did evaluate human reference to get an upper rating but not to contrast two texts (separate HITS)}
%not as much difference as we hoped, basically no difference
%comparing to 3 days when we had the best settings depending on the day the conclusion would be different 

\paragraph{Asking workers to rate both human-written and model-generated stories side-by-side improves ratings:} We hypothesize that the previous result is due to scale calibration differences between the two settings: when repeatedly confronted with incoherent model-generated text, a worker may be more generous with their ratings compared to if they only see coherent human-written text. Thus, we explore whether their ratings can be better calibrated by asking them to rate both types of stories side-by-side, using the same qualification settings as for the other experiments.
The results of this experiment are presented in the middle row of \autoref{tab:AMT_and_teach_hum_gpt2}. Workers score GPT-2 generated stories significantly lower than reference stories on \textit{coherence} (M=3.39 vs. M=3.83), \textit{relevance} (M=2.70 vs. M=3.49), and \textit{likability} (M=2.99 vs. M=3.48), which is in line with our expectations. Their ratings for \textit{grammar} (M=3.82 vs. M=3.83) are similar for both types of text, which we also observe with expert teacher ratings in \autoref{sec:teacher_experiments} and is expected since GPT-2's output is generally grammatical.
% \micomment{add one sentence about agreement?}

%improves results, probably allows better calibration of scores. 

%improved agreement for human written text (relevance) probably due to the fact that it was so much better than gpt2 

%cleared distinction even with people still spamming the task (plus workers get at least 2 texts to evaluate)

% ------------------------------
% Subsection to discuss agreement?
% Relatively low but still better for coherence, grammar, and relevance THAN for likability

% While it may be better with guidelines likability is still highly subjective
% which questions the whole idea of evaluating it in the first place.

%% file: sections/teacher_experiments.tex
\section{Evaluation by expert teachers}
\label{sec:teacher_experiments}

The experiments in the previous section demonstrate the unreliability of AMT ratings for open-ended text generation, even when qualifications are used to restrict the task to ostensibly reliable workers. In this section, we compare the ratings produced by AMT workers to those of expert raters, specifically a set of three English teachers, and discover significant deviations between the two groups. Though they rated both types of stories separately, their ratings clearly distinguish between human-written references and machine-generated stories. We also conducted post-task interviews with the teachers and organized a mediation session to discuss stories with high disagreement, observing that they reach consensus after discussion in about 80\% of cases. 

\paragraph{Recruiting English teachers:}We choose English teachers as experts for our story generation task because they regularly evaluate student-written papers and are experienced at detecting both low-level grammatical mistakes as well as discourse-level issues with logical coherence. The three teachers were recruited from the authors' personal networks, and each of them either has a degree in teaching English as a Second Language or a CELTA certificate.\footnote{\label{CELTA}Certificate in Teaching English to Speakers of Other Languages.} They were paid \$125 each for participating in our experiments, which required them to rate the same 200 human-written stories and 200 GPT-2 generated stories on the same four properties as that of the AMT workers, given an identical task interface.\footnote{The teachers completed their ratings using an identical version of our task deployed on the AMT sandbox environment.}

\paragraph{Unlike AMT workers, teachers rate reference stories higher than GPT-2 generated ones:} 
We asked teachers to first rate the 200 reference stories, and then a week later to rate the GPT-2 generated stories. Just like the AMT workers, they were not told that the text in the second task was machine-generated. Importantly, we used the same set of teachers for both tasks, so they already had significant experience with the task  when rating the machine-generated text (as opposed to using new AMT workers for each experiment).
% This experiment is thus more similar to the AMT task design with paired human-written / model-generated stories described in the previous section.
% \mkcomment{I don't agree with this statement. They do have more experience but the design is not really similar. Having seen human text did NOT help them rate gpt2 output later, if anything it made it more difficult. I would delete this sentence.}

The results of this evaluation are presented in the last row of \autoref{tab:AMT_and_teach_hum_gpt2}. Unsurprisingly, teachers rated human-written stories significantly higher than GPT-2 generated stories in terms of \textit{coherence} (M=4.38 vs. M=3.73), \textit{relevance} (M=3.82 vs. M=2.54), and \textit{likability} (M=3.69 vs. M=2.96) (all \textit{p}'s<0.001). On the other hand, they rated human-written stories and GPT-2 generated stories as similar in terms of \textit{grammar} (M=4.50 vs. M=4.56). 
% Ultimately, these results support the conclusion from the previous section: including human-written references makes it easier to calibrate ratings for both AMT workers and expert raters \mkcomment{this is not true, I actually got comments that it would have been easier if they were together, but it was challenging since the GPT2 output was separate, there was nothing easy in calibrating ratings for GPT2 as they already had it calibrated to human written text (the worst human text they saw was 1 and many gpt2 texts were worse) they all said that it would really help their struggle IF the texts for same prompt were presented together but the way they got them made it more difficult. If anything, one of them told me that it would be better to have GPT2 first, it would be rated low anyway but that would increased ratings for human text (but best to have both together)}. 
Moreover, teachers' ratings of human-written stories are considerably higher than AMT ratings for all attributes except \textit{likability} (M=3.69) which depending on the day was rated lower (M$_{\mathrm{Day 1}}$=3.37) or higher (M$_{\mathrm{Day 2}}$=3.73) by the AMT workers. Similarly, teachers' ratings of GPT-2 stories are lower than the ratings we obtained from AMT workers for \textit{coherence} (M=3.73 vs. M=4.11), \textit{relevance} (M=2.54 vs. M=3.71), and \textit{likability} (M=2.96 vs. M=3.37).

\paragraph{Teachers need to see many examples to properly calibrate their ratings:} 
In post-task interviews, all teachers reported that it took them 10-20 stories on average to calibrate their ratings. Since most AMT workers complete only one to two HITs, they do not have similar time to get acquainted with the task; this may suggest that having a pre-task training phase can improve worker calibration. 

\paragraph{Coherence is difficult to rate for machine-generated text:}
The teachers unanimously report that while \textit{coherence} is easy to rate for reference stories (since most of them are largely coherent), it is the most difficult property to rate for GPT-2 generated stories. Since they did not know that they were rating machine-generated text, they spent time trying to make sense of the author's possible intent in producing many of the strange artifacts and hallucinations common to output of neural language models~\citep{holtzman2019curious}. In contrast, \textit{relevance} turned out to be the easiest property of machine-generated text for teachers to rate, which is expected as many of GPT-2's stories deviate very quickly from the prompt (see \autoref{fig:stories_examples}).

\paragraph{GPT-2 generated stories are much harder for teachers to rate overall:}
All teachers reported struggling more when rating GPT-2 stories, a fact reflected in their average rating time per story increasing significantly from 69.8 seconds to 87.3 seconds (\textit{p}<0.05). In contrast, the average rating time of AMT workers \emph{decreased} from 135.3 seconds for human-written text (Day 1) to 91.5 seconds for GPT-2 text (\textit{p}<0.05)\footnote{This time was computed by the researchers to account for workers accepting multiple HITs at the same time, however, the \textit{WorkTimeInSeconds} reported in the AMT results shows similar trends.}. Teachers also reported having to recalibrate their scale when rating the GPT-2 generated stories, as the stories were significantly worse than the human-written text. Consequently, they suggested that it would be easier to calibrate their scale had the GPT-2 output been presented beside the human-written text, which supports the results from our joint rating task with AMT workers. 
%Furthermore, they also mentioned that in case of GPT-2 generated stories, the previous story affected ratings of the following story, which was not the case in terms of the human-written text. For instance, if the preceding story was of a very poor quality the following story was more likely to be rated higher \mkcomment{updated, agree with Nader's comment} \micomment{in what way?} \nacomment{i feel the previous two sentences: Furthermore... and For instance... can safely be removed if we need space} 
Finally, the teachers suggested that creating a standardized rubric would greatly facilitate the rating process. This step is even more important as machine-generated text faces different issues than human-written text.

%rubric type stuff, 
%gpt2 way more challenging and much slower (+20sec on ave), 
%some ratings were subjective [not sure where to add this ]
%human written stories - coherence easiest to rate, grammar most difficult to calibrate
%gpt2 - relevance easiest to rate, coherence challenging
%would be better to present both stories together with same prompt

\paragraph{Resolving teacher disagreement:} 
One advantage of using human expert raters is that we can easily have them discuss examples on which they disagree. 
We arranged a mediation meeting between two of the three teachers to discuss 60 stories on which they showed the highest disagreement (3 attributes $\times$ 10 stories $\times$ 2 types, we excluded \emph{likability} due to its subjective nature). In this meeting, they were first asked to rate the stories again, without being provided their previous rating. In about 20\% of cases, one of the teachers disagreed with their own previous rating due to honest lapses of judgment. Another common reason for disagreement was missing world knowledge (see \autoref{fig:stories_examples}, right). One more reason for disagreement, a confusion about how to rate slang in terms of \textit{grammaticality}. While the text was not correct in the view of the official grammar, it was appropriate for the prompt, so one teacher rated it high while the other rated it low.
Overall, after discussing examples that they still disagreed on after re-rating, teachers were able to come to a consensus on 80\% of the stories; the remaining disagreements persisted due to individual differences in strictness. See \autoref{sec:appendixc_teachers} for details on the mediation meeting.

\paragraph{Replicating the study on Upwork:}
We recognize that replicating our study is difficult without access to a network of English teachers. As such, we performed the same experiment using three certified teachers recruited on a freelance platform, Upwork.\footnote{\url{https://www.upwork.com}} The teachers were paid \$175 for evaluating the same 200 human-written and 200 GPT-2 generated stories using the exact same setup as in \autoref{sec:crowd_experimental_setup}. It took approximately one week to collect the data (including break between rating human-written and GPT-2 generated stories). The results obtained via Upwork were comparable with the results obtained from the English teachers described in this section, i.e. the Upwork teachers rated human-written stories higher for \textit{coherence}, \textit{relevance}, and \textit{likability} than the GPT-2 generated stories (all \textit{p}'s<0.001). Interestingly, their IAA was higher than the English teachers recruited from the authors' personal networks. The details of this experiment are provided in the \autoref{sec:appendixb_upwork}.

%% file: sections/related_work.tex
\section{Related Work}
\label{sec:related}
Our work is related to previous studies of human evaluation of text quality as well as collecting judgments using Amazon Mechanical Turk.

\paragraph{Human evaluation of text quality:}
Most previous studies on human evaluation concentrate on constrained generation domains, such as machine translation \citep{Guzmn2015, graham_baldwin_moffat_zobel_2017, toral2018attaining, castilho-2021-towards} or summarization \citep{gillick-liu-2010-non, iskender-etal-2020-best}. Other studies evaluate very short, often one sentence long, outputs \citep{Grundkiewicz2015HumanEO, mori-etal-2019-toward, khashabi2021genie}.

Even professional translators struggle when evaluating longer machine translated texts \citep{castilho-2021-towards}. Creative texts, such as stories, are less constrained than translated texts, but researchers continue to employ crowd workers to evaluate creative texts, often without evaluating reference texts (see \autoref{sec:survey}).
% Furthermore, model generated text is often presented in isolation without the reference text (e.g., only 23 out of 45 papers surveyed in \autoref{sec:survey} included reference text). 
Previous studies have asked workers to choose from
\citep{mori-etal-2019-toward} or distinguish between human-written and machine-generated texts \citep{garbacea-etal-2019-judge, ippolito-etal-2020-automatic, clark-etal-2021-thats}.

% \mkcomment{I'm not sure whether this below shouldn't just be commented out as they also had different mturk qual settings}
% \textcolor{red}{\cite{khashabi2021genie} build ``Genie,`` a human-in-the-loop leaderboard, and experimented with collecting ratings across three days. The researchers did not ob
% reported consistent results over three different days of evaluation, in their experiment the workers were evaluating single sentences on only one attribute, which requires arguably less expertise than evaluation of a story passage on multiple attributes.}

%build Genie 

%Our work contributes to the ongoing discussion on human evaluation of generated text. 

% Our work relates to prior analyses of human text generation evaluations. \citet{mori-etal-2019-toward} 
% asked workers %who were evaluating story endings 
% to additionally explain their ratings. \citet{goldfarb-tarrant-etal-2020-content} require workers to also rate how confident they were about their ratings.  
% \citet{novikova-etal-2018-rankme} combine relative rankings and magnitude estimation to improve  interannotator agreement. Other studies filter out  low-quality workers through qualification tests   ~\citep{alva-manchego-etal-2020-asset, iskender-etal-2021-reliability} or by including items with known ratings~\citep{donahue-etal-2020-enabling}. 

\paragraph{Data collection using AMT:} Many previous works raise concerns about the reliability of data collected on AMT~\citep{Necka2016_mturk_research, Matherly2019_mturk_research, Ahler_mturk_dataquality}. Reluctance of requesters to reject HITs leads to positive bias in workers' qualifications \citep{Matherly2019_mturk_research}. Furthermore, a large number of responses are provided by small number of productive workers \citep{Fort2011_prod_workers, Robinson2019_superworkers}. Researchers also report an increasing number of workers use VPNs to mask their location \citep{Bauer2020_mtruk_vpn} and contribute lower-quality data \citep{moss_litman_mturk_vpnlowquality, Ahler_mturk_dataquality}. Hence, simple quality control measures, such as approval rate or the country of residence as suggested in \cite{Berinsky2012mturk_approvalrate}, may not be sufficient to effectively filter workers who are spamming a task.

%% file: sections/recommendations.tex
\section{Recommendations \& Conclusion}
\label{sec:recommendations}

Our experiments show that evaluating open-ended generated text is an incredibly challenging task even for expert raters. While AMT is a convenient and affordable solution, we observe that high variance between workers, poor calibration, and cognitively-demanding tasks can lead researchers to draw misleading scientific conclusions (e.g., that human-written text is ``worse'' than GPT-2's). Simple fixes such as adding strict worker qualifications do not address the root of the problem. As such, we recommend future AMT evaluations implement additional quality control mechanisms (some of which require custom task setups on external servers) such as (1) filtering workers by observed time spent per HIT rather than \textit{WorkTimeInSeconds}, (2) specifying a maximum number of items per worker, (3) employing a pre-task language proficiency test, and (4) providing training HITs to allow workers to calibrate their ratings.  
Furthermore, we show that researchers can improve rating calibration by presenting machine-generated text alongside human reference text. That said, expert raters such as linguists or language teachers should be used whenever possible as they have already been trained to evaluate written text, and it is not much more expensive (it cost us \$144 to rate 200 stories with AMT vs. \$187.50 with English teachers vs. \$262.5 with Upwork. 
% Moreover, it is easier to arrange a mediation meeting in order to resolve differences, identify possible mistakes, discuss the meaning of each score, and get feedback to improve the rating process.

% not black box, explore data, do time-filtering, check for Eng knowledge
% better higher 3 eng teachers/linguists, organize mediation meeting

% %
% - include texts with known rating

% better not to use XCC-> while it may appear convenient and affordable
% would need to: make Eng test, design your interface, check IPs, do quality control and still expect lower quality

%% file: sections/ethics.tex
\section{Ethical Considerations}
\label{sec:ethics}

As with all research that makes use of human subjects, we must carefully reflect on our methodology to minimize the risk of harm to those we ask to evaluate open-ended texts. Specifically, texts from social media sites like Reddit may contain racist, sexist, and other forms of vulgar content. Additionally, neural language models like GPT-2, which have been trained on open domain text crawled from the web, have been shown to generate similarly offensive content. As such, we advocate adequately warning any humans who take part in open-ended text evaluation of the potential for such harms (as we did in our research).

Additionally, crowd workers are frequently underpaid for their labor, which harms both the quality of the research, and more importantly, the ability of these crowd workers to earn an adequate living. As such, we report our hourly wage for both crowd workers and experts. We ensure that crowd workers earn at least \$14 per hour by assuming 50--55 seconds per HIT (though on average our crowd workers were paid substantially higher due to the low average time to completion on each HIT). Our experts averaged around \$20 per hour (not counting mediation).

%% file: sections/acknowledgments.tex
\section*{Acknowledgments}

We thank the reviewers for their insightful comments. We would also like to thank the UMass NLP group for the great advice on the draft of this paper. We are grateful to the English teachers recruited from Upwork and from the authors` personal network, as well as workers on AMT, for their help in the story evaluation. MK is supported by the Chan Zuckerberg Initiative under the project Scientific Knowledge Base Construction. NA and MI are supported by award IIS-1955567 from the National Science Foundation (NSF).

%% file: sections/appendixa.tex
\section{Questions Used for the Survey}
\label{sec:appendixa_survey}

The paper survey described in \autoref{sec:survey} included the following questions: 
\begin{itemize}
    \item type of the task
    \item length of rated text
    \item rated attributes
    \item rating scale
    \item labels used for ratings
    \item definitions of each attribute
    \item instructions provided to raters
    \item qualifications and quality control measures employed
    \item number of rated items
    \item number of rated systems
    \item number of raters per item
    \item inclusion of ground truth
    \item monetary compensation
%    \item type of statistical analysis employed
\end{itemize}

\begin{table*}[!htbp]
\resizebox{\textwidth}{!}{\begin{tabular}{|l|l|l|l|}
\hline
\multicolumn{1}{|c|}{\textbf{Authors}} & \textbf{Year} & \textbf{Title} & \textbf{Venue} \\ \hline
\citeauthor{akoury-etal-2020-storium}   & \citeyear{akoury-etal-2020-storium}       & STORIUM: A Dataset and Evaluation Platform for Machine-in-the-Loop Story Generation         &   EMNLP             \\ \hline
\citeauthor{alshomary-etal-2020-target} & \citeyear{alshomary-etal-2020-target} & Target Inference in Argument Conclusion Generation & ACL \\ \hline
\citeauthor{bosselut-etal-2018-discourse} & \citeyear{bosselut-etal-2018-discourse}  & Discourse-Aware Neural Rewards for Coherent Text Generation & EMNLP\\ \hline
\citeauthor{brahman-chaturvedi-2020-modeling} & \citeyear{brahman-chaturvedi-2020-modeling} & Modeling Protagonist Emotions for Emotion-Aware Storytelling & EMNLP \\ \hline
\citeauthor{chakrabarty-etal-2020-generating} & \citeyear{chakrabarty-etal-2020-generating} & Generating similes effortlessly like a Pro: A Style Transfer Approach for Simile Generation & EMNLP\\ \hline
\citeauthor{chakrabarty-etal-2020-r}  & \citeyear{chakrabarty-etal-2020-r} & R3: Reverse, Retrieve, and Rank for Sarcasm Generation with Commonsense Knowledge & ACL \\ \hline
\citeauthor{clark-etal-2018-neural} & \citeyear{clark-etal-2018-neural} & Neural Text Generation in Stories Using Entity Representations as Context & NAACL\\ \hline
\citeauthor{donahue-etal-2020-enabling} & \citeyear{donahue-etal-2020-enabling} & Enabling Language Models to Fill in the Blanks & ACL \\ \hline
\citeauthor{fan2018hierarchical} & \citeyear{fan2018hierarchical} & Hierarchical Neural Story Generation & ACL \\ \hline
\citeauthor{fang-etal-2020-video2commonsense} & \citeyear{fang-etal-2020-video2commonsense} & Video2Commonsense: Generating Commonsense Descriptions to Enrich Video Captioning & EMNLP\\ \hline
\citeauthor{goldfarb-tarrant-etal-2020-content} & \citeyear{goldfarb-tarrant-etal-2020-content} & Content Planning for Neural Story Generation with Aristotelian Rescoring & EMNLP\\ \hline
\citeauthor{gorinski-lapata-2018-whats} & \citeyear{gorinski-lapata-2018-whats} & What`s this Movie about? A Joint Neural Network Architecture for Movie Content Analysis & NAACL \\ \hline
\citeauthor{goyal-durrett-2020-neural} & \citeyear{goyal-durrett-2020-neural} & Neural Syntactic Preordering for Controlled Paraphrase Generation & ACL \\ \hline
\citeauthor{he-etal-2019-pun} & \citeyear{he-etal-2019-pun} & Pun Generation with Surprise & NAACL\\ \hline
\citeauthor{hegel-etal-2020-substance} & \citeyear{hegel-etal-2020-substance} & Substance over Style: Document-Level Targeted Content Transfer & EMNLP \\ \hline
\citeauthor{holtzman-etal-2018-learning}  & \citeyear{holtzman-etal-2018-learning} & Learning to Write with Cooperative Discriminators & ACL\\ \hline
\citeauthor{hsu-etal-2019-visual} & \citeyear{hsu-etal-2019-visual} & Visual Story Post-Editing & ACL\\ \hline
\citeauthor{ippolito-etal-2019-unsupervised} & \citeyear{ippolito-etal-2019-unsupervised} & Unsupervised Hierarchical Story Infilling & NAACL \\ \hline
\citeauthor{jiang-etal-2020-neural} & \citeyear{jiang-etal-2020-neural} & Neural CRF Model for Sentence Alignment in Text Simplification & ACL \\ \hline
\citeauthor{krishna-etal-2020-reformulating} & \citeyear{krishna-etal-2020-reformulating} & Reformulating Unsupervised Style Transfer as Paraphrase Generation & EMNLP \\ \hline
\citeauthor{kriz-etal-2019-complexity} & \citeyear{kriz-etal-2019-complexity} & Complexity-Weighted Loss and Diverse Reranking for Sentence Simplification & NAACL\\ \hline
\citeauthor{li-etal-2018-delete} & \citeyear{li-etal-2018-delete} & Delete, Retrieve, Generate: A Simple Approach to Sentiment and Style Transfer & NAACL \\ \hline
\citeauthor{lin-etal-2020-learning} & \citeyear{lin-etal-2020-learning} & Learning to Generate Multiple Style Transfer Outputs for an Input Sentence & ACL \\ \hline
\citeauthor{liu-etal-2019-towards-explainable} & \citeyear{liu-etal-2019-towards-explainable} & Towards Explainable NLP: A Generative Explanation Framework for Text Classification & ACL \\ \hline
\citeauthor{mallinson-etal-2020-zero} & \citeyear{mallinson-etal-2020-zero} & Zero-Shot Crosslingual Sentence Simplification & EMNLP\\ \hline
\citeauthor{martins-etal-2020-sparse} & \citeyear{martins-etal-2020-sparse} & Sparse Text Generation & EMNLP \\ \hline
\citeauthor{mir-etal-2019-evaluating} & \citeyear{mir-etal-2019-evaluating} & Evaluating Style Transfer for Text & NAACL \\ \hline
\citeauthor{peng-etal-2018-towards} & \citeyear{peng-etal-2018-towards} & Towards Controllable Story Generation & NAACL \\ \hline
\citeauthor{pezeshkpour-etal-2018-embedding} & \citeyear{pezeshkpour-etal-2018-embedding} & Embedding Multimodal Relational Data for Knowledge Base Completion & EMNLP\\ \hline
\citeauthor{qin-etal-2020-back} & \citeyear{qin-etal-2020-back} & Embedding Multimodal Relational Data for Knowledge Base Completion & EMNLP \\ \hline
\citeauthor{qin-etal-2019-counterfactual} & \citeyear{qin-etal-2019-counterfactual} & Counterfactual Story Reasoning and Generation & EMNLP \\ \hline
\citeauthor{rao-tetreault-2018-dear} & \citeyear{rao-tetreault-2018-dear} & Dear Sir or Madam, May I Introduce the GYAFC Dataset: Corpus, Benchmarks and Metrics for Formality Style Transfer & NAACL \\ \hline
\citeauthor{rashkin-etal-2020-plotmachines} & \citeyear{rashkin-etal-2020-plotmachines} & PLOT MACHINES: Outline-Conditioned Generation with Dynamic Plot State Tracking & EMNLP\\ \hline
\citeauthor{shen-etal-2019-towards} & \citeyear{shen-etal-2019-towards} & Towards Generating Long and Coherent Text with Multi-Level Latent Variable Models & ACL \\ \hline
\citeauthor{sudhakar-etal-2019-transforming} & \citeyear{sudhakar-etal-2019-transforming} & ``Transforming'' Delete, Retrieve, Generate Approach for Controlled Text Style Transfer & EMNLP \\ \hline
\citeauthor{tu-etal-2019-generating} & \citeyear{tu-etal-2019-generating} & Generating Diverse Story Continuations with Controllable Semantics & EMNLP \\ \hline
\citeauthor{tu-etal-2019-generating} & \citeyear{tu-etal-2019-generating} & Can Humor Prediction Datasets be used for Humor Generation? Humorous Headline Generation via Style Transfer & ACL \\ \hline
\citeauthor{xu-etal-2020-megatron} & \citeyear{xu-etal-2020-megatron} & MEGATRON - CNTRL : Controllable Story Generation with External Knowledge Using Large-Scale Language Models & EMNLP \\ \hline
\citeauthor{yang-etal-2019-end-end} & \citeyear{yang-etal-2019-end-end} & An End-to-End Generative Architecture for Paraphrase Generation & EMNLP \\ \hline
\citeauthor{modi-parde-2019-steep} & \citeyear{modi-parde-2019-steep} & The Steep Road to Happily Ever After: An Analysis of Current Visual Storytelling Models & NAACL \\ \hline
\citeauthor{yu-wan-2019-avoid} & \citeyear{yu-wan-2019-avoid} & How to Avoid Sentences Spelling Boring? Towards a Neural Approach to Unsupervised Metaphor Generation & NAACL \\ \hline
\citeauthor{yu-etal-2018-neural} & \citeyear{yu-etal-2018-neural} & A Neural Approach to Pun Generation & ACL \\ \hline
\citeauthor{yu-etal-2020-routing} & \citeyear{yu-etal-2020-routing} & Routing Enforced Generative Model for Recipe Generation & EMNLP \\ \hline
\citeauthor{zhang-tetreault-2019-email} & \citeyear{zhang-tetreault-2019-email} & This Email Could Save Your Life: Introducing the Task of Email Subject Line Generation & ACL \\ \hline
\citeauthor{zang-etal-2019-automated} & \citeyear{zang-etal-2019-automated} & Automated Chess Commentator Powered by Neural Chess Engine & ACL \\ \hline
\end{tabular}}
\caption{List of Surveyed Papers.}
\label{tab:surveyed_papers}
\end{table*}

%% file: sections/appendixb.tex
\newpage
\section{Collecting Ratings on Upwork}
\label{sec:appendixb_upwork}

We also hired three teachers using the freelancing platform Upwork\footnote{\url{https://www.upwork.com/}}. The teachers were paid \$175 to evaluate the same 200 human-written stories and 200 GPT-2 generated stories. They were asked to perform the ratings on the AMT platform in order to use the same interface as workers on AMT. Similarly to the teachers recruited from the authors' personal network, the teachers recruited on Upwork  were asked to rate the 200 human-written stories first and then, after a few days break, provide the ratings for the GPT-2 generated stories. Furthermore, Upwork teachers also held TEFL,\footnote{Teaching English as a Foreign Language.} TESOL,\footnote{Teaching English to Speakers of Other Languages} or CELTA\footnotemark[\getrefnumber{CELTA}] certificates.  \autoref{tab:upwork_story_ratings} shows mean ratings and agreement for the data collected on Upwork. Similarly to the results described in \autoref{sec:teacher_experiments} and summarized in \autoref{tab:teachers_ttest}, the average scores for \textit{coherence}, \textit{relevance}, and \textit{likability} are higher for the human-written stories than for the GPT-2 generated stories (see \autoref{tab:upwork_ttest}).

\begin{table*}[ht]

    \small
    \centering
    \begin{tabular}{lllllllll}
    \toprule
    \multirow{2}*{\textbf{Experiment description}}  & \multicolumn{2}{c}{\hspace*{-3ex}\underline{\textbf{Grammar}}} & \multicolumn{2}{c}{\hspace*{-3ex}\underline{\textbf{Coherence}}} & \multicolumn{2}{c}{\hspace*{-3ex}\underline{\textbf{Relevance}}} & \multicolumn{2}{c}{\hspace*{-3ex}\underline{\textbf{Likability}}} \\
    & Mean$_{\text{\emph{STD}}}$ & IAA$_{\text{\emph{\%}}}$ & Mean$_{\text{\emph{STD}}}$ & IAA$_{\text{\emph{\%}}}$ & Mean$_{\text{\emph{STD}}}$ & IAA$_{\text{\emph{\%}}}$ & Mean$_{\text{\emph{STD}}}$ & IAA$_{\text{\emph{\%}}}$ \\
    \midrule
   % & \multicolumn{8}{c}{\emph{Ratings of Upwork Teachers}} \vspace{0.15cm}\\
    Reference & 4.28$_{{0.78}}$ & 0.29$_{{26}}$ & 4.55$_{{0.66}}$ & 0.11$_{{30.5}}$ & 
    4.25$_{{0.88}}$ & 0.49$_{{26.4}}$ & 4.02$_{{1.16}}$ & 0.01$_{{14.5}}$ \\
    
    GPT-2 & 4.25$_{{0.82}}$ & 0.12$_{{19}}$ & 3.99$_{{0.98}}$ & 0.06$_{{12.5}}$ & 3.02$_{{1.53}}$ & 0.25$_{{5}}$ & 3.68$_{{1.15}}$ & 0.05$_{{10.5}}$ \\ 

    \bottomrule
    \end{tabular}
    \caption{Ratings of human-written reference stories and GPT-2 generated stories collected on \textbf{Upwork}. Inter-annotator agreement (IAA) between the three raters is measured with Krippendorff’s $\alpha$ well as the percentage of stories for which all three raters exactly agreed on a rating (the latter is subscripted)}
    \label{tab:upwork_story_ratings}
\end{table*}

\begin{table*}[ht]
\small
\centering
\begin{tabular}{llllllllll}
  \hline
 & mean (human) & mean (GPT-2) & difference & 95\% CI lower & 95\% CI upper &  t & df & \textit{p}-val \\ 
  \hline
grammar & 4.28 & 4.25 & 0.03 & -0.06 & 0.12 & 0.72 & 1194.9 & 0.47 \\ 
coherence & 4.55 & 3.99 & 0.56 & 0.46 & 0.65 & 11.46 & 1044.3 & \textbf{<0.001} 
\\ 
relevance & 4.02 & 3.02 & 1.00 & 0.85 & 1.15 & 12.82 & 1103.8 & \textbf{<0.001}  \\ 
likability & 4.23 & 3.83 & 0.40 & 0.30 & 0.53 & 7.24 & 1143.4 & \textbf{<0.001} \\ 
   \hline
\end{tabular}
\caption{Welch`s \textit{t}-test for ratings collected on \textbf{Upwork} (human-written stories vs GPT-2 generated stories). Human-written stories were rated higher on \textit{coherence}, \textit{relevance}, and \textit{likability} than GPT-2 generated stories. These results are similar to the one obtained from English teachers described in \autoref{sec:teacher_experiments}.}
\label{tab:upwork_ttest}
\end{table*}

%% file: sections/appendixc.tex
\section{Details on Post-rating Interviews}
\label{sec:appendixc_teachers}

Two mediation meetings were organized with two of the three teachers (due to availability) over Zoom~\footnote{\url{https://zoom.us/}}. The teachers were asked to reevaluate 60 stories on which they showed disagreement (3 attributes $\times$ 10 stories $\times$ 2 types; \textit{likability} was excluded due to its subjective nature). Each meeting took approximately 2h (including a short break) and was led by one of the authors. The teachers were shown one story at a time and were asked to reevaluate it on the given attribute. In about 20\% of the cases, the teachers agreed with each other, suggesting that the previous disagreement was due to honest lapses of judgment. As for the cases where disagreement occurred, each was asked to provide a justification for their ratings. Often hearing the other party's argument enabled them to see the text from a different perspective and understand the ratings of the other person. This process often resulted in them adjusting their own ratings. Common reasons for disagreement which could be resolved during the mediation meeting included: world knowledge, difference in understanding of the prompt and its relation to the text (e.g., prompt enforcing specific style), difference in the way they treated author's comments which were sometimes present at the beginning of the story, and rationalizing connections between the sentences.

After each batch, consisting of ratings of both human-written stories and GPT-2 generated stories, each of the three teachers took part in a short one-on-one interview ($\sim$ 10min each). They were asked the following questions:

\begin{enumerate}
\item How long did it take you to calibrate your ratings?
\item Explain in more detail how you rated \textit{coherence}/\textit{grammar}/\textit{likability}/\textit{relevance}? What was the process. Did you have to reread the text?
\item How did you calibrate the ratings for \textit{coherence}/\textit{grammar}/\textit{likability}/\textit{relevance}? What constituted a 5? What about a 1?
\item How often did you take breaks?
\item Which attribute was the easiest to calibrate? Which was the most difficult?
\item Any other comments or suggestions?
\end{enumerate}

Additionally, after the second batch of GPT-2 stories, the teachers were also asked: (1) which batch was better written, (2) which batch included more computer generated stories and to what extent, (3) whether they had to recalibrate their ratings, and (4) whether they would prefer to see both batches at the same time.

%% file: sections/appendixd.tex
\afterpage{\clearpage}
\section{Statistical Analysis}
\label{sec:appendixd_mturk_results}

\begin{table*}[ht]
\small
\centering
\begin{tabular}{llllllllll}
  \hline
 & mean (human) & mean (GPT-2) & difference & 95\% CI lower & 95\% CI upper &  t & df & \textit{p}-val \\ 
  \hline
grammar & 4.50 & 4.55 & 0.05 & -0.14 & 0.03 & -1.27 & 1111.7 & 0.21 \\ 
coherence & 4.38 & 3.73 & 0.65 & 0.53 & 0.77 & 10.63 & 1119.6 & \textbf{<0.001} \\ 
relevance & 3.82 & 2.54 & 1.28 & 1.12 & 1.44 & 15.45 & 1190.7 & \textbf{<0.001}  \\ 
likability & 3.69 & 2.96 & 0.73 & 0.57 & 0.89 & 9.16 & 1182 & \textbf{<0.001}\\ 
   \hline
\end{tabular}
\caption{Welch`s \textit{t}-test for ratings collected in the experiment described in \autoref{sec:teacher_experiments} (teachers' ratings). Human-written stories were rated higher for \textit{coherence}, \textit{relevance}, and \textit{likability} than GPT-2 generated stories.}
\label{tab:teachers_ttest}
\end{table*}

\begin{figure*}[ht]
    \centering
    \includegraphics[width=\textwidth,height=\textheight,keepaspectratio]{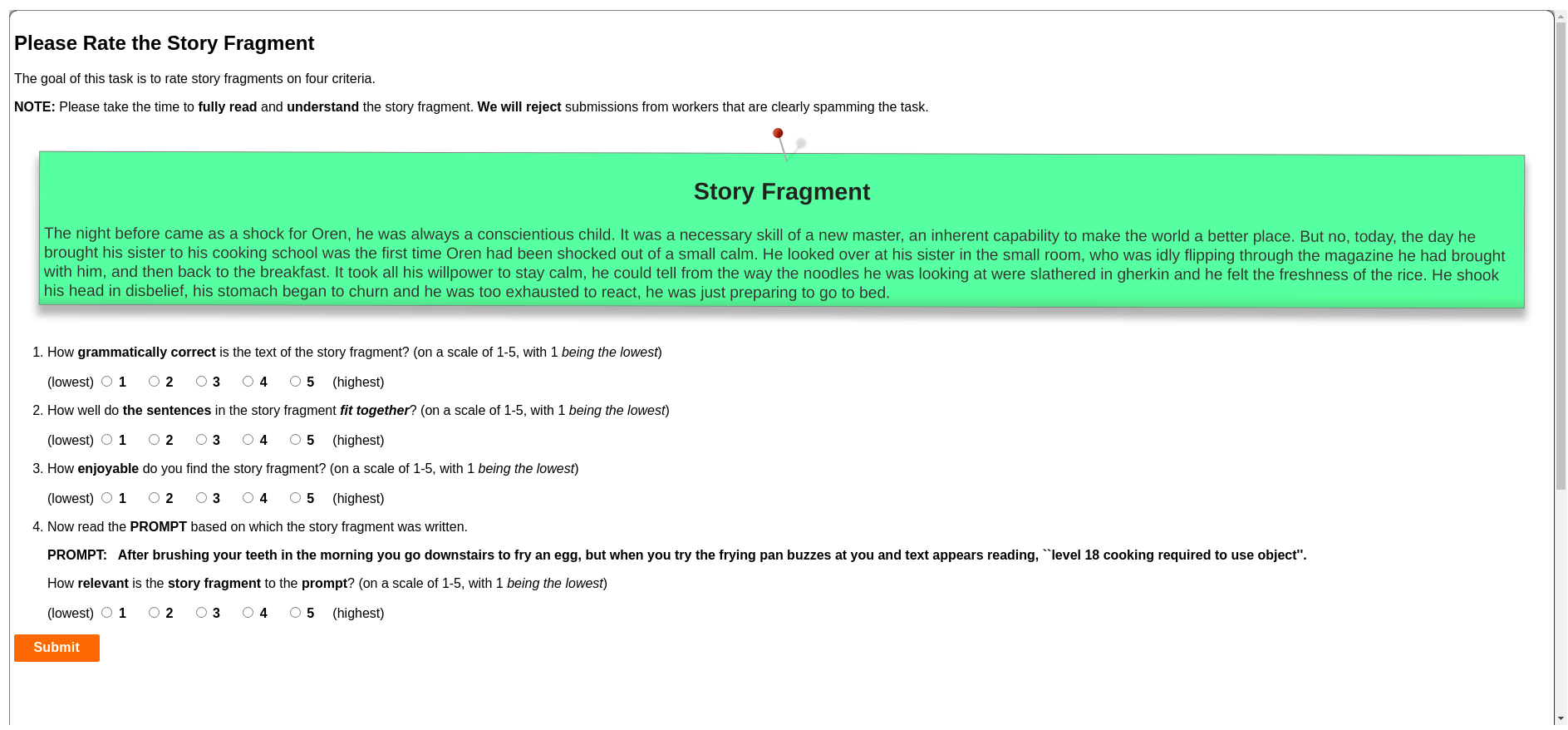}
    \caption{AMT interface for evaluation of one story.}
    \label{fig:one_story_eval}
\end{figure*}

\begin{figure*}[ht]
    \centering
    \includegraphics[width=\textwidth,height=\textheight,keepaspectratio]{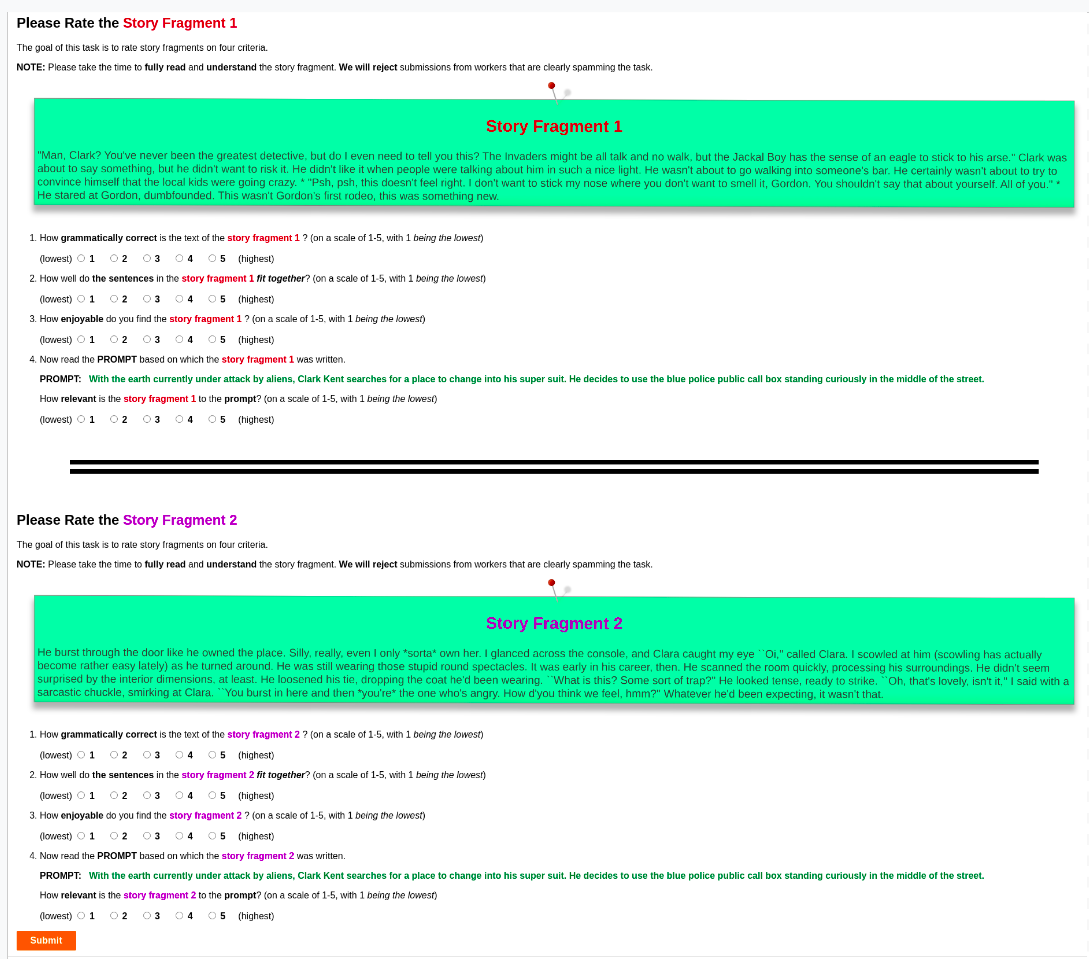}
    \caption{AMT interface for evaluation of both types of stories (GPT-2 and human reference).}
    \label{fig:two_stories_eval}
\end{figure*}

%η2 = 0.01 indicates a small effect; η2 = 0.06 indicates a medium effect; η2 = 0.14 indicates a large effect.

%GRAMMAR DAYS1-3 and NNS
% latex table generated in R 3.6.3 by xtable 1.8-4 package
\begin{table*}[ht]
\centering
\begin{tabular}{lrrrrrr}
  \hline
 & Df & Sum Sq & Mean Sq & F value & Pr($>$F) & $\eta_{p}^{2}$ \\ 
  \hline
  Group & 3 & 13.63 & 4.54 & 4.90 & \textbf{0.002} & 0.01 \\ 
  Residuals & 2396 & 2223.51 & 0.93 &  &  \\ 
   \hline
\end{tabular}
\caption{One-way ANOVA investigating the effect of group (Day 1, Day 2, Day 3, and workers from non-English-speaking countries) on the ratings of \textit{\textbf{grammar}} of the reference texts. Partial eta squared ($\eta_{p}^{2}$) is provided for the effect size ($\eta_{p}^{2}$ = 0.01 indicates small effect size; $\eta_{p}^{2}$ = 0.06 indicates medium effect size;  $\eta_{p}^{2}$ = 0.14 indicates large effect size \citep{cohen:statistical1988}).}
\end{table*}

%POST-HOC DAYs 1-3 and NNS for GRAMMAR
% latex table generated in R 3.6.3 by xtable 1.8-4 package
\begin{table*}[ht]
\centering
\begin{tabular}{rrrr}
  \hline
 & Day 1 & Day 2 & Day 3 \\ 
  \hline
  Day 2 & 0.08 &  &  \\ 
  Day 3 & 1.00 & 0.20 &  \\ 
  NNS & \textbf{0.01} & 1.00 & \textbf{0.03} \\ 
   \hline
\end{tabular}
\caption{Pairwise post-hoc test with Bonferroni adjustment for the ratings of \textit{\textbf{grammar}} between Day 1, Day 2, Day 3, and non-English speaking countries (NNS). The numbers provided in the table are \textit{p}-values for the given pairwise comparison. Grammar ratings provided by the workers from non-English speaking countries are significantly different from ratings provided by the workers from English-speaking countries on Day 1 and Day 3~\footnote{We observed lower number of workers with unreasonably short completion time for these two days when compared with Day 2.}  } 
\end{table*}

%%%%%%%%%%%%%%%%%%%%%%%%%%%%%%%%%%%%%%%%%%%%%%%%%%%%%%%%%%%%%%

%COHERENCE - BETWEEN MTURK DATA
% latex table generated in R 3.6.3 by xtable 1.8-4 package
%Shows ANOVA for Day1-Day3 and NNS for COHERENCE 
\begin{table*}[ht]
\centering
\begin{tabular}{lrrrrrr}
  \hline
 & Df & Sum Sq & Mean Sq & F value & Pr($>$F) ) & $\eta_{p}^{2}$ \\ 
  \hline
Group & 3 & 161.28 & 53.76 & 51.35 & \textbf{<0.001} & 0.06 \\ 
  Residuals & 2396 & 2508.21 & 1.05 &  &  \\ 
   \hline
\end{tabular}
\caption{One-way ANOVA investigating the effect of group (Day 1, Day 2, Day 3, and workers from non-English-speaking countries) on the ratings of \textit{\textbf{coherence}} of the reference texts. Partial eta squared ($\eta_{p}^{2}$) is provided for the effect size ($\eta_{p}^{2}$ = 0.01 indicates small effect size;  $\eta_{p}^{2}$ = 0.06 indicates medium effect size;  $\eta_{p}^{2}$ = 0.14 indicates large effect size \citep{cohen:statistical1988}).}
\end{table*}

%post-hoc for ANOVA Days1-3 + NNS for COHERENCE
% latex table generated in R 3.6.3 by xtable 1.8-4 package
\begin{table*}[ht]
\centering
\begin{tabular}{rrrr}
  \hline
 & Day 1 & Day 2 & Day 3 \\ 
  \hline
Day 2 & 1.00 &  &  \\ 
  Day 3 & \textbf{<0.001} & \textbf{<0.001} &  \\ 
  NNS & \textbf{<0.001} & \textbf{<0.001} & \textbf{0.02} \\ 
   \hline
\end{tabular}
\caption{Pairwise post hoc test with Bonferroni adjustment for the ratings of \textit{\textbf{coherence}} between Day 1, Day 2, Day 3, and non-English speaking countries (NNS). The numbers provided in the table are \textit{p}-values for the given pairwise comparison. Ratings of \textit{coherence} provided by raters from non-English speaking countries are significantly different from ratings of workers from English-speaking countries. Furthermore, there are some difference between Day 1, Day 2, and Day 2.} 
\end{table*}

%%%%%%%%%%%%%%%%%%%%%%%%%%%%%%%%%%%%%%%%%%%%%%%%%
%RELEVANCE - ANOVA for Days1-3 and NNS
% latex table generated in R 3.6.3 by xtable 1.8-4 package
\begin{table*}[ht]
\centering
\begin{tabular}{lrrrrrrr}
  \hline
 & Df & Sum Sq & Mean Sq & F value & Pr($>$F) &  $\eta_{p}^{2}$ \\ 
  \hline
Group & 3 & 89.99 & 30.00 & 19.92 & \textbf{<0.001} & 0.02 \\ 
  Residuals & 2396 & 3607.85 & 1.51 &  & & \\ 
   \hline
\end{tabular}
\caption{One-way ANOVA investigating the effect of group (Day 1, Day 2, Day 3, and workers from non-English-speaking countries) on the ratings of \textbf{\textit{relevance}} of the reference texts. Partial eta squared ( $\eta_{p}^{2}$) is provided for the effect size ($\eta_{p}^{2}$ = 0.01 indicates small effect size;  $\eta_{p}^{2}$ = 0.06 indicates medium effect size;  $\eta_{p}^{2}$ = 0.14 indicates large effect size \citep{cohen:statistical1988}).}
\end{table*}

%RELEVANCE -- Days1-3 NNS 
% latex table generated in R 3.6.3 by xtable 1.8-4 package
\begin{table*}[ht]
\centering
\begin{tabular}{rrrr}
  \hline
 & Day 1 & Day 2 & Day 3 \\ 
  \hline
  Day 2 & \textbf{0.01} &  &  \\ 
  Day 3 & 1.00 & 0.14 &  \\ 
  NNS & \textbf{<0.001} & \textbf{<0.001} & \textbf{<0.001} \\ 
   \hline
\end{tabular}
\caption{Pairwise post hoc test with Bonferroni adjustment for the ratings of \textit{\textbf{relevance}} between Day 1, Day 2, Day 3, and non-English speaking countries (NNS). The numbers provided in the table are \textit{p}-values for the given pairwise comparison. Ratings obtained from workers from non-English speaking countries differ significantly from ratings obtained from workers from English-speaking countries on Day 1, Day 2, and Day 3. Furthermore, there is a significant difference between ratings collected on Day 1 and Day 2.} 
\end{table*}

%DAYS 1-3 and NNS for LIKABILITY 
% latex table generated in R 3.6.3 by xtable 1.8-4 package
\begin{table*}[ht]
\centering
\begin{tabular}{lrrrrrr}
  \hline
 & Df & Sum Sq & Mean Sq & F value & Pr($>$F)  &  $\eta_{p}^{2}$  \\ 
  \hline
Group & 3 & 62.70 & 20.90 & 15.89 & \textbf{<0.001} & 0.02 \\ 
  Residuals & 2396 & 3151.22 & 1.32 &  &  & \\ 
   \hline
\end{tabular}
\caption{One-way ANOVA investigating the effect of group (Day 1, Day 2, Day 3, and workers from non-English-speaking countries) on the ratings of \textit{\textbf{likability}} of the reference texts. Partial eta squared ($\eta_{p}^{2}$) is provided for the effect size ($\eta_{p}^{2}$ = 0.01 indicates small effect size;  $\eta_{p}^{2}$ = 0.06 indicates medium effect size;  $\eta_{p}^{2}$ = 0.14 indicates large effect size \citep{cohen:statistical1988}).}
\end{table*}

% latex table generated in R 3.6.3 by xtable 1.8-4 package
% post-hoc likability bonferroni DAYS 1-3 and NNS
\begin{table*}[ht]
\centering
\begin{tabular}{rrrr}
  \hline
 & Day 1 & Day 2 & Day 3 \\ 
  \hline
  Day 2 & \textbf{<0.001} &  &  \\ 
  Day 3 & 1.00 & \textbf{<0.001} &  \\ 
  NNS & 1.00 & \textbf{<0.001} & 0.75 \\ 
   \hline
\end{tabular}
\caption{Pairwise post hoc test with Bonferroni adjustment for the ratings of \textit{\textbf{likability}} between Day 1, Day 2, Day 3, and non-English speaking countries (NNS). The numbers provided in the table are \textit{p}-values for the given pairwise comparison. Ratings provided by workers from non-English speaking countries differ significantly from ratings obtained from workers from English-speaking countries on Day 2. Furthermore, there are significant differences between ratings obtained on Day 1 and Day 2, as well as between ratings obtained on Day 2 and Day 3.} 
\end{table*}

%Welch t-test for Day 1-3 and NNS for 4 attributes
\begin{table*}[ht]
\small
\centering
\begin{tabular}{cccccccccc}
  \hline
 & mean (human) & mean (GPT-2) & difference & 95\% CI lower & 95\% CI upper &  t & df & \textit{p}-val \\ 
  \hline
grammar & 4.00 & 3.94 & 0.06 & -0.05 & 0.16 & 1.06 & 1197.9 & 0.29 \\ 
coherence & 4.11 & 3.82 & 0.29 & 0.18 & 0.42 & 4.97 & 1169.1 & \textbf{<0.001} \\ 
relevance & 3.71 & 3.44 & 0.27 & 0.12 & 0.43 & 3.54 & 1184.7 & \textbf{<0.001}  \\ 
likability & 3.37 & 3.42 & 0.05 & -0.18 & 0.09 & -0.64 & 1194.5 & 0.52 \\ 
   \hline
\end{tabular}
\caption{Welch`s \textit{t}-test on ratings collected on AMT for human-written stories (Day 1) and GPT-2 generated stories. Human-written stories are being rated higher for \textit{coherence} and more \textit{relevance} than GPT-2 generated stories (\textit{p}<0.05).}
\end{table*}

\begin{table*}[ht]
\small
\centering
\begin{tabular}{cccccccccc}
  \hline
 & mean (human) & mean (GPT-2) & difference & 95\% CI lower & 95\% CI upper &  t & df & \textit{p}-val \\ 
  \hline
grammar & 3.86 & 3.94 & 0.08 & -0.19 & 0.03 & -1.50 & 1197.9 & 0.14 \\ 
coherence & 3.92 & 3.82 & 0.10 & -0.02 & 0.23 & 1.72 & 1176.5 & 0.09 \\ 
relevance & 3.71 & 3.44 & 0.27 & 0.13 & 0.41 & 3.69 & 1123.5 & \textbf{<0.001}  \\ 
likability & 3.73 & 3.42 & 0.31 & 0.19 & 0.44 & 4.89 & 1128.9 & \textbf{<0.001} \\ 
   \hline
\end{tabular}
\caption{Welch`s \textit{t}-test on ratings collected on AMT for human-written stories (Day 2) and GPT-2 generated stories. Human-written stories were rated higher for \textit{relevance} and \textit{likability} than GPT-2 generated stories (\textit{p}<0.05).}
\end{table*}

\begin{table*}[ht]
\small
\centering
\begin{tabular}{cccccccccc}
  \hline
 & mean (human) & mean (GPT-2) & difference & 95\% CI lower & 95\% CI upper &  t & df & \textit{p}-val \\ 
  \hline
grammar & 3.98 & 3.94 & 0.04 & -0.07 & 0.15 & 0.70 & 1196.4 & 0.48 \\ 
coherence & 4.05 & 3.82 & 0.23 & 0.12 & 0.36 & 3.98 & 1163.5 & \textbf{<0.001} \\ 
relevance & 3.46 & 3.44 & 0.02 & -0.13 & 0.17 & 0.27 & 1188.9 & 0.80 \\ 
likability & 3.42 & 3.42 & 0.00 & -0.14 & 0.14 & 0.00 & 1192.5 & 1 \\ 
   \hline
\end{tabular}
\caption{Welch`s \textit{t}-test for ratings collected on AMT for human-written stories (Day 3) and GPT-2 generated stories. Human-written stories were rated higher for \textit{coherence} than GPT-2 generated stories (\textit{p}<0.05).}
\end{table*}

\begin{table*}[ht]
\small
\centering
\begin{tabular}{cccccccccc}
  \hline
 & mean (human) & mean (GPT-2) & difference & 95\% CI lower & 95\% CI upper &  t & df & \textit{p}-val \\ 
  \hline
grammar & 3.82 & 3.94 & 0.12 & -0.23 & -0.01 & -2.11 & 1183.1 & \textbf{0.04} \\ 
coherence & 3.45 & 3.82 & 0.37 & -0.49 & -0.23 & -5.42 & 1194.2 & \textbf{<0.00}1 \\ 
relevance & 3.25 & 3.44 & 0.19 & -0.35 & -0.04 & -2.50 & 1185 & \textbf{0.01}  \\ 
likability & 3.32 & 3.42 & 0.10 & -0.24 & 0.04 & -1.41 & 1197.9 & 0.16 \\ 
   \hline
\end{tabular}
\caption{Welch`s \textit{t}-test for ratings collected on AMT for human-written stories (non-English speaking countries) and GPT-2 generated stories. GPT-2 generated stories were rated \textit{higher} for \textit{grammar}, \textit{coherence}, and \textit{relevance} than human-written stories (\textit{p}<0.05)}
\end{table*}

\begin{table*}[ht]
\small
\centering
\begin{tabular}{cccccccccc}
  \hline
 & mean (human) & mean (GPT-2) & difference & 95\% CI lower & 95\% CI upper &  t & df & \textit{p}-val \\ 
  \hline
grammar & 3.83 & 3.82 & 0.01 & -0.09 & 0.12 & 0.28 & 1188.1 & 0.78 \\ 
coherence & 3.83 & 3.39 & 0.44 & 0.32 & 0.57 & 6.92 & 1198 & \textbf{<0.001} \\ 
relevance & 3.49 & 2.70 & 0.79 & 0.65 & 0.93 & 10.85 & 1198 & \textbf{<0.001}  \\ 
likability & 3.48 & 2.99 & 0.49 & 0.37 & 0.62 & 7.72 & 1195.2 & \textbf{<0.001} \\ 
   \hline
\end{tabular}
\caption{Welch`s \textit{t}-test for ratings collected on AMT for human-written stories and GPT-2 generated stories (both stories shown in one HIT). GPT-2 generated stories were rated lower for \textit{coherence}, \textit{relevance}, and \textit{likability} than human-written stories (\textit{p}<0.05) which is in line with the ratings provided by English teachers.}
\end{table*}